\documentclass[a4paper]{article}
\usepackage{cite}
\usepackage{amsmath,amssymb,amsfonts}
\usepackage{algorithmic}
\usepackage{graphicx}
\usepackage{textcomp}
\usepackage{authblk}
\usepackage[square,numbers]{natbib}

\usepackage{url}
\usepackage{doi}
\usepackage{multirow}%
\usepackage{amsthm}%
\usepackage{mathrsfs}%
\usepackage{manyfoot}%
\usepackage{booktabs}%
\usepackage{listings}%
\usepackage{footmisc}
\usepackage{bm}

\usepackage{tikz}
\usetikzlibrary{arrows, arrows.meta, calc, positioning, decorations.pathreplacing, shapes.geometric}

\tikzstyle{nn}=[circle,thick,draw=black!75,minimum size=6mm,fill=white]
\tikzstyle{rr}=[rectangle,rounded corners,thick,draw=black!75,minimum size=6mm,fill=white]
\usepackage{hyperref} 
\hypersetup{
    bookmarksnumbered=true, bookmarksopen=true,
	unicode=true, colorlinks=true, linktoc=all, 
	linkcolor=blue, citecolor=blue, filecolor=blue, urlcolor=blue,
	pdfstartview=FitH
}

\RequirePackage{tabularx} 						    
\newcolumntype{L}{>{$}l<{$}}                        
\newcolumntype{R}{>{$}r<{$}}                        
\newcolumntype{Y}{>{\centering\arraybackslash}X}    
\newcolumntype{Z}{>{\raggedleft\arraybackslash}X}   

\usepackage{placeins}

\providecommand{\keywords}[1]{\textbf{\textit{Keywords---}} #1}

\begin{document}

\title{Whole-Graph Representation Learning \\ For the Classification of Signed Networks}
\author[1]{Noé Cécillon}
\author[1]{Vincent Labatut}
\author[2]{Richard Dufour}
\author[3]{Nejat Arinik}
\affil[1]{LIA UPR 4128, Avignon Université, Avignon, F-84911, France (e-mail: \texttt{\{firstname.lastname\}@univ-avignon.fr})}
\affil[2]{LS2N UMR 6004, Nantes Université, Nantes, F-44322, France (e-mail: \texttt{richard.dufour@univ-nantes.fr})}
\affil[3]{CRIL UMR 8188, Université d'Artois, Lens, F-62307, France (e-mail: \texttt{arinik@cril.fr})}

\maketitle

\begin{abstract}
Graphs are ubiquitous for modeling complex systems involving structured data and relationships. Consequently, graph representation learning, which aims to automatically learn low-dimensional representations of graphs, has drawn a lot of attention in recent years. The overwhelming majority of existing methods handle \textit{unsigned} graphs. However, \textit{signed} graphs appear in an increasing number of application domains to model systems involving two types of opposed relationships. Several authors took an interest in signed graphs and proposed methods for providing \textit{vertex}-level representations, but only one exists for \textit{whole-graph} representations, and it can handle only \textit{fully connected} graphs. 
In this article, we tackle this issue by proposing two approaches to learning \textit{whole-graph} representations of general \textit{signed} graphs. The first is a SG2V, a signed generalization of the whole-graph embedding method Graph2vec that relies on a modification of the Weisfeiler--Lehman relabelling procedure. The second one is WSGCN, a whole-graph generalization of the signed vertex embedding method SGCN that relies on the introduction of master nodes into the GCN. We propose several variants of both these approaches. A bottleneck in the development of whole-graph-oriented methods is the lack of data. We constitute a benchmark composed of three collections of signed graphs with corresponding ground truths. We assess our methods on this benchmark, and our results show that the signed whole-graph methods learn better representations for this task. Overall, the baseline obtains an $F$-measure score of $58.57$, when SG2V and WSGCN reach $73.01$ and $81.20$, respectively. Our source code and benchmark are publicly available online. 
\end{abstract}


\section{Introduction}
Graph representation learning is a general task consisting of automatically learning a data-driven, low-dimensional and fixed-size vector representation of graphs, or parts of graphs such as vertices, edges, and subgraphs, that preserves the information conveyed by their structure~\cite{Hamilton2020}. The main benefit of such representations is that one can feed them to standard machine learning tools, and thus process graphs indirectly with general and efficient methods, instead of designing task-specific methods required to handle graphs directly. Due to this advantage over traditional feature engineering methods, the development of graph representation learning approaches has been the object of many publications in the last few years, as attested by the many recent surveys on the topic~\cite{Zhang2019ae, Kim2021a, Liu2022ai}. Most existing methods focus on vertices or edges, which is very useful in several tasks such as community detection or link prediction. However, the representation of graphs \textit{as a whole} is also crucial, especially in graph classification~\cite{Tsuda2010, Morris2022}. This task typically involves a collection of graphs, each one belonging to a specific class to be predicted.

In addition, the vast majority of existing representation learning methods are meant to handle \textit{unsigned} graphs. This is understandable, as the number of available unsigned graph datasets is much larger than that for signed graphs. Nevertheless, signed graphs appear in many application domains such as Sociology~\cite{Neal2022}, Neurosciences~\cite{Tang2022}, International Relations~\cite{Doreian2015}, Business Science~\cite{Jensen2006}, Finance~\cite{MacMahon2013}, Political Science~\cite{Doreian2019}, and Computer Science~\cite{Arinik2021a}; and their processing requires appropriate tools. Such graphs were originally introduced in Psychology, to represent the attitudes of people toward other people or objects~\cite{Heider1946}. More generally, they can be used to model any system involving two types of semantically opposed relationships (like/dislike, similar/different, etc.). In general, this duality makes it impossible to directly apply standard unsigned methods. Several methods have been proposed recently to handle signed graph representation learning at the level of \textit{vertices}~\cite{Yuan2017, Kim2018b, Derr2018}. However, to the best of our knowledge, only one approach~\cite{Tang2022} allows learning \textit{whole-graph} representations of signed graphs, and it is limited to \textit{fully connected} graphs (i.e., every vertex is connected to all other vertices). As a consequence, it cannot handle most of the real-world signed networks from the literature, which are typically sparse (e.g., \cite{Facchetti2012, Brito-Montes2022}). 

In this paper, we tackle this issue by proposing two approaches to learning the whole-graph representation of signed networks. The first is a signed generalization of the unsigned whole-graph embedding method \textit{Graph2vec}~\cite{Narayanan2017}. The second one is a whole-graph generalization of the signed vertex embedding method Signed Graph Convolutional Network (SGCN)~\cite{Derr2018}. Our contributions are threefold: 
\begin{itemize}
    \item The first is methodological and concerns the two proposed representation learning approaches, for which we define several variants. All are able to handle not only fully connected graphs, but also sparse ones. 
    \item The second is resource-oriented, as we constitute and share a benchmark annotated for signed graph classification, and constituted of three distinct collections. This is a first, as all similar datasets only focus on unsigned graphs. 
    \item The third contribution is experimental, as we apply our proposed methods to our benchmark to assess their performance. Our results show that the proposed approaches perform better than the baseline. 
\end{itemize}

The rest of this article is organized as follows. In Section~\ref{sec:Background}, we introduce the main concepts and notations used later and review existing methods for graph representation learning. Next, we present our signed graph classification benchmark, in Section~\ref{sec:Datasets}. Then, in Section~\ref{sec:Methods}, we describe the methods that we propose to handle signed whole-graph representation learning. We present and compare our results in Section~\ref{sec:Results}. Finally, we summarize our main findings in Section~\ref{sec:Conclusion}, and discuss possible perspectives.


\section{Background}
\label{sec:Background}
This section describes the main notions used in the rest of the article. We first introduce the principal concepts and notations related to signed graphs in Section~\ref{sec:BackSignedGraphs}). We then review existing methods for graph representation learning in Section~\ref{sec:BackWholeGrEmb}.

\subsection{Signed Graphs}
\label{sec:BackSignedGraphs}
Formally, a signed graph is a triple $G = (V,E,s)$ composed of a set of vertices $V$, a set of edges $E \subseteq V^2$ between them, and a function $s: E \rightarrow \{-,+\}$ that associates a sign to each edge. We denote $E^-$ and $E^+$ as the subsets of negative and positive edges, respectively. Consequently, $E = E^- \cup E^+$. As is customary in graph theory~\cite{Bollobas1998}, we denote $n$ as the \textit{order} of the graph (i.e., its number of vertices), while $m$ is its \textit{size} (i.e., its number of edges). 

In this work, we focus on undirected unweighted signed graphs. The \textit{unsigned} neighborhood $N(u)$ of a vertex $u$ ignores edge signs and includes all vertices attached to this vertex: $N(u) = \{ v \in V : (u,v) \in E \}$. On the contrary, the positive $N^+(u)$ and negative $N^-(u)$ neighborhoods focus only on one edge sign: $N^\pm(u) = \{ v \in V : (u,v) \in E^\pm \}$. We similarly define the unsigned, negative and positive degrees as the cardinalities of the corresponding neighborhoods, i.e., $k(u) = |N(u)|$, $k^-(u) = |N^-(u)|$, and $k^+(u) = |N^+(u)|$. The sign of a path or cycle corresponds to the product of its constituting edge signs. Consequently, this sign is negative if the path or cycle contains an odd number of negative edges, and positive otherwise. The positive (resp. negative) \textit{reachable set} of a vertex $u$ is the subset of vertices that are connected to $u$ through positive (resp. negative) shortest paths. 

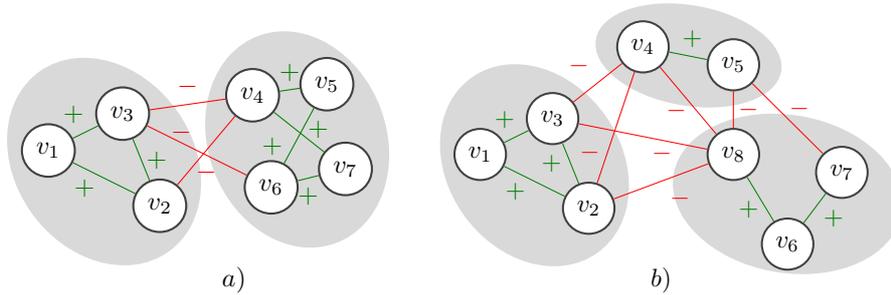
\begin{figure}[!t]
    \centering
    \resizebox{0.45\linewidth}{!}{\begin{tikzpicture}
    \fill[fill=gray!30,rotate around={-55:(6.75,-0.15)}] (6.75,-0.15) ellipse (1.5cm and 1.2cm);
    \fill[fill=gray!30,rotate around={20:(9.35,0.15)}] (9.35,0.15) ellipse (1.2cm and 1.5cm);
    
    \node [nn] (v1) at ( 6.00,  0.00) {$v_1$};
    \node [nn] (v2) at ( 7.50, -0.75) {$v_2$};
    \node [nn] (v3) at ( 7.00,  0.50) {$v_3$};
    \node [nn] (v4) at ( 8.75,  0.75) {$v_4$};
    \node [nn] (v5) at ( 9.75,  0.90) {$v_5$};
    \node [nn] (v6) at ( 9.00, -0.50) {$v_6$};
    \node [nn] (v7) at (10.00, -0.25) {$v_7$};

    \draw[color=green!50!black] (v1) edge node[below left=-0.1cm and -0.0cm] {$+$} (v2);
    \draw[color=green!50!black] (v1) edge node[above left=0.0cm and -0.1cm] {$+$} (v3);
    \draw[color=green!50!black] (v2) edge node[right=-0.05cm] {$+$} (v3);
    \draw[color=red] (v2) edge node[below=0.05cm] {$-$} (v4);
    \draw[color=red] (v3) edge node[above] {$-$} (v4);
    \draw[color=red] (v3) edge node[above left=0.0cm and -0.05cm] {$-$} (v6);
    \draw[color=green!50!black] (v4) edge node[above] {$+$} (v5);
    \draw[color=green!50!black] (v4) edge node[above right=-0.2cm and 0.0cm] {$+$} (v7);
    \draw[color=green!50!black] (v5) edge node[below left=-0.1cm and 0.1cm] {$+$} (v6);
    \draw[color=green!50!black] (v6) edge node[below] {$+$} (v7);
    
    \node at (8.50,-1.75) {$a)$};
\end{tikzpicture}}
    \hfill
    \resizebox{0.53\linewidth}{!}{\begin{tikzpicture}
    \fill[fill=gray!30,rotate around={-55:(6.75,-0.15)}] (6.75,-0.15) ellipse (1.5cm and 1.2cm);
    \fill[fill=gray!30,rotate around={-10:(8.88,1.38)}] (8.88,1.38) ellipse (1.3cm and 0.7cm);
    \fill[fill=gray!30,rotate around={-10:(10.25,-0.55)}] (10.25,-0.55) ellipse (1.6cm and 1.1cm);

    \node [nn] (v1) at ( 6.00,  0.00) {$v_1$};
    \node [nn] (v2) at ( 7.50, -0.75) {$v_2$};
    \node [nn] (v3) at ( 7.00,  0.50) {$v_3$};
    \node [nn] (v4) at ( 8.25,  1.50) {$v_4$};
    \node [nn] (v5) at ( 9.50,  1.25) {$v_5$};
    \node [nn] (v6) at (10.25, -1.25) {$v_6$};
    \node [nn] (v7) at (11.00, -0.25) {$v_7$};
    \node [nn] (v8) at ( 9.50,  0.00) {$v_8$};

    \draw[color=green!50!black] (v1) edge node[below left=-0.1cm and -0.0cm] {$+$} (v2);
    \draw[color=green!50!black] (v1) edge node[above left=0.0cm and -0.1cm] {$+$} (v3);
    \draw[color=green!50!black] (v2) edge node[left=0.00cm] {$+$} (v3);
    \draw[color=green!50!black] (v4) edge node[above right=0.0cm and -0.2cm] {$+$}  (v5);
    \draw[color=green!50!black] (v6) edge node[below right=-0.1cm and 0.0cm] {$+$}  (v7);
    \draw[color=green!50!black] (v6) edge node[below left=-0.1cm and -0.1cm] {$+$}  (v8);
    \draw[color=red] (v2) edge node[below left=0.1cm and 0.1cm] {$-$}  (v4);
    \draw[color=red] (v2) edge node[below right=0.0cm and 0.0cm] {$-$}  (v8);
    \draw[color=red] (v3) edge node[above left=0.0cm and 0.0cm] {$-$}  (v4);
    \draw[color=red] (v3) edge node[below right=0.0cm and 0.0cm] {$-$}  (v8);
    \draw[color=red] (v4) edge node[below left=-0.1cm and -0.1cm] {$-$}  (v8);
    \draw[color=red] (v5) edge node[right=-0.05cm] {$-$}  (v8);
    \draw[color=red] (v5) edge node[above right=-0.1cm and -0.1cm] {$-$}  (v7);
    
    \node at (8.50,-1.75) {$b)$};
\end{tikzpicture}}
    \caption{Examples of perfectly balanced graphs according to a) Structural Balance; and b) Generalized Balance ($k=3$).}
    \label{fig:StructBalance}
\end{figure}

\textit{Structural Balance} (SB) is a fundamental property of signed graphs~\cite{Heider1946, Harary1953}. 
In its strict definition, a graph is said to be structurally balanced when all its cycles are positive~\cite{Cartwright1956}. Equivalently, for a structurally balanced graph, it is possible to find a bisection of $V$ such that all positive edges are internal, i.e., they connect vertices from the same cluster, whereas all negative edges are external, i.e., they lie in between clusters. Fig.~\ref{fig:StructBalance}.a) illustrates this situation: the graph contains two clusters $\{v_1, v_2, v_3 \}$ and $\{v_4, ..., v_7 \}$; all positive edges are inside these clusters; all negative edges are between them. In real-world networks, though, graphs are rarely \textit{perfectly} balanced, and no bisection exists that respects the SB definition. In this case, one may want to measure the \textit{amount} of imbalance in the graph. This is typically done by computing the \textit{Frustration} measure (a.k.a. \textit{Line Index} or \textit{Imbalance}), which requires solving a combinatorial optimization problem~\cite{Doreian1996}. Let us consider an arbitrary bisection of $V$. The positive edges located in between clusters and the negative edges located inside them are said to be \textit{frustrated}, as they do not respect SB. For instance, in Fig.~\ref{fig:StructBalance}.a) , if edges $(v_1, v_2)$ and $(v_2, v_4)$  were negative and positive, respectively, they would be frustrated. The Frustration of this bisection is the number of such edges. The Frustration of the graph is the minimal Frustration over all possible bisections. Put differently, the graph Frustration is the minimal number of edges whose sign must be switched to reach perfect SB.

The notion of SB was generalized to allow partitions composed of more than two antagonistic clusters~\cite{Davis1967}. A graph that can be split into $k$ such clusters is said to be $k$-balanced. Equivalently, this \textit{Generalized Balance} (GB) requires that a graph have no cycles with \textit{exactly one} negative edge. Fig.~\ref{fig:StructBalance}.b) illustrates the notion of GB: the presented graph contains three clusters $\{v_1, v_2, v_3\}$, $\{v_4, v_5\}$, and $\{v_6, ..., v_8\}$; all positive edges are inside the clusters; all negative edges are between them. Note that Frustration remains a valid imbalance measure for this generalization of the SB. The combinatorial problem consisting of finding the partition of $V$ that minimizes Frustration is called the \textit{Correlation Clustering} (CC) problem~\cite{Bansal2002}. 
Finally, it is important to stress that the ubiquity of SB among real-world systems, which is widely advertised in the literature, is nevertheless disputed among specialists of signed graphs~\cite{Doreian2017}. Therefore, it is not guaranteed that building a graph representation method that takes SB into account will perform better than one that ignores it. Assessing this point is one of the goals of our experiments.

\subsection{Graph Representation Learning}
\label{sec:BackWholeGrEmb}
Graph representation learning is a very popular research topic, and the literature provides a profusion of methods allowing us to automatically train models to represent various \textit{types} of graphs (directed, weighted, signed, multiplex, etc.), and various \textit{parts} of graphs (vertices, edges, subgraphs, whole graphs, etc.). However, to the best of our knowledge, none of them can handle \textit{signed} graphs as a \textit{whole}, except Hierarchical Signed Graph Representation Learning (HSGRL)~\cite{Tang2022}. Moreover, this method, which we discuss later in this section, suffers from a limitation as it can only process \textit{fully connected} graphs. 
Consequently, to position our work, in the following, we summarize the main approaches designed to deal with one of two situations that are the closest to our own: on the one hand, representing \textit{whole} but \textit{unsigned} graphs, and on the other hand, representing the \textit{vertices} of \textit{signed} graphs. The methods used as a basis for our work are described in further detail later, in Section~\ref{sec:Methods}.

\subsubsection{Whole Graphs}
Whole-graph embeddings can be obtained through handcrafted approaches such as feature engineering~\cite{Papegnies2019}, matrix factorization~\cite{Maddalena2021}, or graph kernels~\cite{Kriege2020}. However, they often suffer from generalization problems~\cite{Narayanan2017}. Representation learning methods are data-driven and allow solving this issue by adapting the graph representation to the targeted dataset. 

A popular approach involves leveraging models from the field of Natural Language Processing (NLP), such as in Graph2vec~\cite{Narayanan2017} and Graph Classification via Graph Structure Learning (GC-GSL)~\cite{Huynh2022}. Graph2vec, one of the earliest methods for whole-graph representation learning, is an unsupervised and task-agnostic approach that leverages the Doc2vec model~\cite{Le2014} from NLP and applies it to graphs. Instead of treating a text as a collection of paragraphs, Graph2vec considers a graph as a collection of subgraphs, which are then used to train a SkipGram model (cf. Section~\ref{sec:MethodsSGCN} for more details).

Another family of approaches relies on graph autoencoders to learn the representation in an unsupervised way, e.g., Permutation-Invariant Graph-level Autoencoder (PIGAE)~\cite{Winter2021}, or the Denoising Autoencoder-based (DAE) method from~\cite{Gutierrez-Gomez2019}. Such neural networks are composed of two parts. First, the encoder receives a raw representation of the graph, which is compressed to remove redundant information and superfluous variability and get a fixed-sized and compact representation. Second, the decoder is in charge of reconstructing the original input based on the compressed representation. The autoencoder is trained by minimizing the reconstruction error.


The literature contains another family of approaches, that adapt Convolutional Neural Networks (CNN) from the field of image processing to handle graphs, resulting in supervised methods able to learn whole-graph representations for specific classification tasks, e.g., Patchy-San Convolutional Network (PSCN)~\cite{Niepert2016} and NgramCNN~\cite{Luo2017a}. PSCN adapts the notion of convolution to the context of graphs, which allows applying the same principle as for image processing. The graph is represented by a collection of subgraphs, on which PSCN performs convolutions. These are then aggregated to create higher-level representations of the graph.


Finally, another strategy is to modify vertex-oriented Graph Neural Networks (GNN) to produce whole-graph representations, e.g., Message Passing
Neural Network (MPNN)~\cite{Gilmer2017} or Virtual Column Network (VCN)~\cite{Pham2017}. This is conducted by the addition of a so-called master node, which is connected to all the other vertices. At the end of the training, the vector associated with this master node can be used as a representation of the graph. 

\subsubsection{Signed Graphs}
Although much less common than for \textit{unsigned} graphs, many \textit{vertex} representation learning methods have been developed for \textit{signed} graphs. Historically, the first type of such method relied on random walks, e.g., Signed Network Embedding (SNE)~\cite{Yuan2017} or Signed Directed Embeddings (SIDE)~\cite{Kim2018b}. The general idea is to sample the graph using random walks and feed them to a standard neural network, which learns a representation that preserves both graph structure and edge signs. Put differently, the representations of two well-connected vertices tend to be close in the embedding space, whereas those of two vertices connected by a negative edge tend to be distant. 

Signed Network Embedding (SiNE)~\cite{Wang2017} relies on a deep learning framework. Unlike the previous methods, it does not use random walks to decompose the network, but simpler subgraphs as it extracts all the open triads. The neural network is trained to learn a representation of a vertex that is similar to its direct positive neighbors, and dissimilar to its direct negative neighbors. We describe SiNE in further detail in Section~\ref{sec:MethodsEmbedding}.

More recently, a group of methods leveraged GNN (Graph Neural Networks) to learn a representation of vertices in signed graphs. The first one, Signed Graph Convolutional Network (SGCN)~\cite{Derr2018}, directly generalizes Graph Convolutional Networks (GCNs)~\cite{Kipf2017} to signed graphs, by proposing a dual representation and a dual message passing rule to take positive and negative paths into account (cf. Section~\ref{sec:MethodsSGCN} for more details). Methods such as Signed Graph Attention Networks (SiGAT)~\cite{Huang2019b} or Signed Network Embedding via Graph Attention (SNEA)~\cite{Li2020c} introduce attention in the process, to give more importance to relevant neighbors during the message passing step.

Hierarchical Signed Graph Representation Learning (HSGRL)~\cite{Tang2022} extends SNEA by including an additional pooling module. Based on an information-based centrality metric, it selects a fixed number of vertices and uses their individual representations to build an overall representation of the \textit{whole} signed graph. According to our review of the literature, HSGRL is the \textit{only} method able to do so. However, it is designed only for \textit{fully connected} graphs: this restricts its application to a very specific type of graphs and constitutes a serious limitation of this method. Indeed, the signed networks used in the literature to model and study real-world systems are typically sparse, see for instance~\cite{Facchetti2012, Brito-Montes2022}.

\section{Signed Graph Datasets}
\label{sec:Datasets}
To assess the proposed methods, we constitute a benchmark composed of three datasets of signed networks. This is not a trivial work, as most publicly available signed networks are \textit{individual} graphs that are used for single graph problems such as vertex classification or link prediction. To perform graph classification, not only do we need \textit{collections} of signed graphs, but these collections must be \textit{annotated} for classification (i.e., each graph must be explicitly associated with a class). Our three datasets come from various sources and differ in their structures and sizes. The first one (Section~\ref{sec:DataSpaceOrigin}) is based on an existing collection of real-world \textit{unsigned} networks originally designed for a binary classification task, which we extend to obtain \textit{signed} networks. The second one (Section~\ref{sec:DataCorrClust}) is an existing collection of artificially generated signed graphs, and the third one (Section~\ref{sec:DataEuroParl}) is an existing collection of real-world vote networks. These last two collections were not initially designed for graph classification: we leverage their metadata to repurpose them and define proper classification tasks. In principle, we could also include the \textit{fully connected} signed networks used to assess the performance of the HSGRL method in~\cite{Tang2022}. Unfortunately, they are not publicly available, therefore, we focus on the three datasets that we constructed. Global statistics describing these datasets are provided in Table~\ref{tab:StatsDatasets}, and described later (Section~\ref{sec:DataStats}). The datasets themselves are all publicly available online\footnote{\url{https://doi.org/10.5281/zenodo.13851362} \label{ftn:Datasets}}.

\subsection{SpaceOrigin Conversations}
\label{sec:DataSpaceOrigin}
The \textit{SpaceOrigin} collection (SO) was originally proposed in~\cite{Papegnies2019}. Papegnies \textit{et al}. extract a collection of conversational networks from a corpus of chat conversations taking place between players of the online video game SpaceOrigin. Each network is built around a message of interest, called the \textit{targeted message}, and aims at modeling its conversational context. Its vertices represent players, and its weighted edges reflect the intensity of their verbal interactions. Each network integrates the messages present in a so-called \textit{context period}, which contains a fixed number of messages occurring right before and after the targeted message. Temporal integration is performed by sliding a fixed-sized window over the context period, and incrementing edge weights based on the co-occurrence of speakers in this window. Papegnies \textit{et al}. 
tackle the task of automatic moderation, which they formulate as a binary classification problem consisting of determining whether the targeted message is \textit{Abusive} or \textit{Non-abusive}. The available ground truth is based on manual annotation. For more details on the graph extraction process and the task itself, see~\cite{Papegnies2019}.

The networks produced in~\cite{Papegnies2019} are unsigned, though. To obtain signed networks instead, we change some parts of the extraction process. When sliding the window over the context period, we leverage a sentiment analyzer\footnote{\url{https://github.com/TheophileBlard/french-sentiment-analysis-with-bert}} to determine the polarity of the players' interactions, based on their exchanged textual content. The resulting weight change can thus be either negative (hostile interaction) or positive (neutral or friendly). Consequently, the total weight obtained when integrating over the whole context period can also be negative or positive. We call the resulting dataset \textit{Signed SpaceOrigin} (SSO), and it contains $2{,}545$ conversational graphs. 

\subsection{Correlation Clustering Instances}
\label{sec:DataCorrClust}
This dataset is proposed in~\cite{Arinik2021}, originally as a means to study the space of optimal solutions to the \textit{Correlation Clustering} problem~\cite{Bansal2002} (CC), described in Section~\ref{sec:BackSignedGraphs}. 
Arınık \textit{et al}. want to study the multiplicity and diversity of the optimal solutions to CC. For this purpose, they define a random model and generate a collection of artificial graphs with planted partitions, applying various levels of noise to control the difficulty of the problem. They use an exact method to identify all possible optimal solutions for each graph in this collection and study how certain graph characteristics relate to the number of solutions. 
Due to the NP-hard nature of CC, they focus on relatively small graphs, with a maximal order (number of vertices) of $n=50$. They produce a total of $24{,}660$ unweighted signed graphs, including $22{,}560$ completely connected graphs (i.e., every pair of vertices is connected), while the remaining $2{,}100$ graphs are not completely connected, with a density ranging from $0.25$ to $0.75$.

To use this dataset in the present work, we define a classification problem by associating each graph of the collection with a label. This problem, named \textit{Correlation Clustering Solutions} (CCS), consists of predicting whether there are a single vs. several optimal CC solutions for the graph of interest. 

\begin{table}[!t]
    \caption{Statistics describing our three datasets. Notations $\pm$ and $[~]$ respectively denote the standard deviation, and minimum \& maximum.}
    \label{tab:StatsDatasets}
    \centering
    \begin{tabularx}{\textwidth}{X r r@{ }l@{ }r r@{ }l@{ }r}
        \hline
        \textbf{Data} & \textbf{Number} & \multicolumn{3}{r}{\textbf{Average Number}} & \multicolumn{3}{r}{\textbf{Average density}} \\
         & \textbf{of Graphs} & \multicolumn{3}{r}{\textbf{of Vertices}} & \multicolumn{3}{r}{\textbf{~}} \\
        \hline
        SSO &  $2{,}545$ & $47.74$ & ±$20.34$ & $[ 2;214]$ & $0.48$ & $\pm 0.16$ & $[0.10;1.00]$ \\
        CCS & $24{,}660$ & $27.31$ &  ±$7.44$ & $[16; 50]$ & $0.95$ & $\pm 0.17$ & $[0.19;1.00]$ \\
        EPF &  $6{,}000$ & $67.34$ & ±$59.21$ & $[20;274]$ & $0.70$ & $\pm 0.19$ & $[0.07;1.00]$ \\
        \hline
    \end{tabularx}

    \medskip
    \begin{tabularx}{\textwidth}{X r r r@{ }l@{ }r r@{ }l@{ }r}
        \hline
        \textbf{Data} & \textbf{Nbr. of} & \textbf{Gini} & \multicolumn{3}{r}{\textbf{Average}} & \multicolumn{3}{r}{\textbf{Average}} \\
         & \textbf{Classes} & \textbf{Index} & \multicolumn{3}{r}{\textbf{SB Frustration}} & \multicolumn{3}{r}{\textbf{GB Frustration}} \\
        \hline
        SSO & $2$ & $0.74$ & $0.30$ & $\pm 0.04$ & $[0.01;0.48]$ & $0.25$ & $\pm 0.04$ & $[0.01;0.46]$ \\
        CCS & $2$ & $0.66$ & $0.37$ & $\pm 0.05$ & $[0.03;0.51]$ & $0.33$ & $\pm 0.04$ & $[0.01;0.49]$ \\
        EPF & $3$ & $0.44$ & $0.28$ & $\pm 0.04$ & $[0.01;0.46]$ & $0.22$ & $\pm 0.03$ & $[0.00;0.45]$ \\
        \hline     
    \end{tabularx}
    
    \medskip
    \begin{tabularx}{\textwidth}{X r@{ }l@{ }r r@{ }l@{ }r r@{ }l@{ }r}
        \hline
        \textbf{Data} & \multicolumn{3}{r}{\textbf{Average Number}} & \multicolumn{3}{r}{\textbf{Average Number}} \\
        & \multicolumn{3}{r}{\textbf{of Negative Edges}} & \multicolumn{3}{r}{\textbf{of Positive Edges}} \\
        \hline
        SSO & $166.1$ &  $\pm 22.96$ & $[ 1; 1{,}692]$ & $    245.9$ &      $\pm 30.41$ & $[ 1; 2{,}323]$ \\
        CCS & $220.6$ & $\pm 130.45$ & $[25;     833]$ & $    131.0$ &      $\pm 80.11$ & $[24;     392]$ \\
        EPF & $333.9$ & $\pm 877.20$ & $[ 0;15{,}933]$ & $2{,}552.2$ & $\pm 5{,}761.46$ & $[ 0;33{,}153]$ \\
        \hline
    \end{tabularx}
    
    \medskip
    \begin{tabularx}{\textwidth}{X r@{ }l@{ }r}
        \hline
        \textbf{Data} & \multicolumn{3}{r}{\textbf{Average proportion}} \\
        & \multicolumn{3}{r}{\textbf{of Positive Edges}} \\
        \hline
        SSO & $59.61$ & $\pm 24.73$ & $[ 0.00;100.00]$ \\
        CCS & $37.54$ & $\pm 11.99$ & $[20.10; 79.01]$ \\
        EPF & $78.30$ & $\pm 22.22$ & $[ 0.00;100.00]$ \\
        \hline
    \end{tabularx}
\end{table}

\subsection{European Parliament Roll-Calls}
\label{sec:DataEuroParl}
The last dataset is based on a collection of signed graphs extracted in~\cite{Arinik2020} from a description of the voting activity at the European Parliament (EP). The raw data corresponds to roll-call votes cast individually by Members of the EP (MEPs) during plenary sessions, in the course of the 7th term (2009--2014). Such votes can take one of three values: \textsc{For} (MEP supporting the proposition), \textsc{Against} (MEP opposing the proposition) or \textsc{Abstention} (MEP not taking a stand despite being present). MEPs can also be absent, and consequently, not take part at all in a roll-call. Each network extracted by Arınık \textit{et al}. corresponds to a specific roll-call, using vertices to model MEPs and edges to represent an agreement between them: a positive sign represents an identical vote, and a negative one represents a disagreement. The goal of Arınık \textit{et al}. is to study the polarization of the EP, and more specifically, its voting patterns, and how these are affected by various criteria, such as the topic of the voted proposition. For this purpose, they first identify factions of similarly voting MEPs in each roll-call network. Next, they compare the resulting vertex partitions to identify the types of situations that result in a comparable voting pattern. 

The raw data contains $6{,}595$ roll calls, from which the Arınık \textit{et al}. extract many more networks by leveraging the metadata associated with the voting activity. Since each MEP belongs to a member state and a European political group, they extract not only overall networks containing all MEPs, but also state- and group-specific networks, containing only the MEPs from a given member state or political group, respectively. In total, they produce $244{,}015$ networks.
Some of them are too small or too sparse (almost empty) to be interesting in a classification context, though. We constitute our dataset by first filtering out these unusable instances and then randomly sampling $6{,}000$ networks.

As for the previous dataset, the initial collection of networks considered here was not originally used to perform any prediction tasks. 
We leverage the clusters of networks exhibiting similar voting patterns identified by Arınık \textit{et al}., and define a classification task consisting of predicting the number of factions identified in a network. We call the resulting dataset \textit{European Parliament Factions} (EPF).

\subsection{Brief Comparison}
\label{sec:DataStats}
Table~\ref{tab:StatsDatasets} provides a few descriptive statistics for all three datasets, to help interpret the classification results in Section~\ref{sec:Results}. For each dataset, the top part shows the number of graphs, the average number of vertices by graph, and the average density. Graph density is the proportion of edges present in the graph, relative to a fully connected graph. CCS is by far the largest dataset in terms of the number of graphs, however, these are smaller. Moreover, the order of graphs, (i.e., their number of vertices) is not as uniform in both SSO and EPF, covering two orders of magnitude. 

The middle part of the table shows the number of classes in each dataset, the Gini Index, and the average Frustration. The datasets contain roughly the same number of classes. The Gini Index is used here to characterize class imbalance. The Frustration measure, which we compute for both types of considered balance (SB vs. GB) is expressed as a proportion over the total number of edges in the graph, to have comparable values. By definition, the Frustration obtained for the strict version of structural balance is greater or equal to that of the generalized version: equal if the optimal partition contains two clusters, and greater if it contains three or more clusters. 

Finally, the bottom part of the table exhibits the average numbers of positive and negative edges by graph, and the average proportion of positive edges. The number of edges is quite variable in all three datasets. In EPF, a few graphs have no positive or negative edges at all. Overall, the graphs tend to be denser than \textit{unsigned} real-world networks~\cite{Esmailian2015}.

\section{Representation Methods}
\label{sec:Methods}
We now describe the three families of methods that we propose to handle signed whole-graph representation learning. The first can be considered a baseline and relies on the aggregation of signed \textit{vertex} embeddings (Section~\ref{sec:MethodsEmbedding}). The second is an adaptation of an \textit{unsigned} whole-graph embedding method to \textit{signed} graphs (Section~\ref{sec:MethodsSG2V}). The third is based on a Graph Convolutional Network able to learn signed \textit{vertex} representations,  which we adapt to handle \textit{whole} signed graphs (Section~\ref{sec:MethodsSGCN}).

\subsection{Aggregated Signed Network Embedding}
\label{sec:MethodsEmbedding}
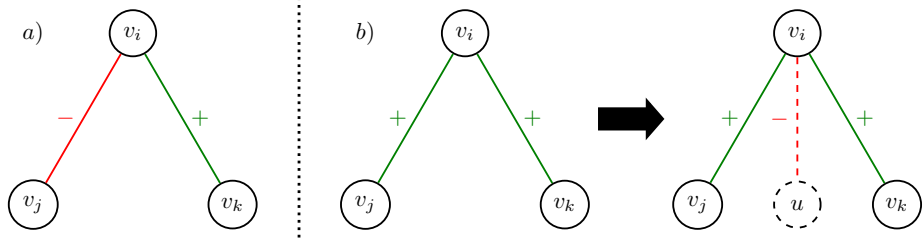
\begin{figure}[!t]
    \centering
    \resizebox{1\linewidth}{!}{\begin{tikzpicture}
    \tikzstyle{nn}=[circle, thick, draw=black, minimum size=7mm, fill=none]
		\node[nn] (v1) at (0, 2.60) {$v_i$};
		\node[nn] (v2) at (-1.5, 0) {$v_j$};
		\node[nn] (v3) at (1.5, 0) {$v_k$};
		\draw[thick, color=red] (v1) edge node[left] {$-$} (v2);
		\draw[thick, color=green!50!black] (v1) edge node[right] {$+$} (v3);
        \node at (-1.5,2.60) {$a)$};

        \draw[very thick, dotted] (2.5, -0.5) -- (2.5, 3.10);
        
        \node[nn] (v4) at (5, 2.60) {$v_i$};
		\node[nn] (v5) at (3.5, 0) {$v_j$};
		\node[nn] (v6) at (6.5, 0) {$v_k$};
		\draw[thick, color=green!50!black] (v4) edge node[left] {$+$} (v5);
		\draw[thick, color=green!50!black] (v4) edge node[right] {$+$} (v6);
        \node at (3.5,2.6) {$b)$};

        \draw[-{Triangle[width=18pt,length=8pt]}, line width=10pt] (7,1.30) -- (8,1.30);
        
        \node[nn] (v7) at (10, 2.60) {$v_i$};
		\node[nn] (v8) at (8.5, 0) {$v_j$};
		\node[nn] (v9) at (11.5, 0) {$v_k$};
		\node[nn, dashed] (v10) at (10.0, 0) {$u$};
		\draw[thick, color=green!50!black] (v7) edge node[left] {$+$} (v8);
		\draw[thick, color=green!50!black] (v7) edge node[right] {$+$} (v9);
		\draw[thick, color=red, dashed] (v7) edge node[left] {$-$} (v10);
\end{tikzpicture}}
    \caption{Triads used in SiNE, as inputs of the dual neural network. Mixed triplets (a) are used directly, positive ones (b) require a transformation, and negative ones (not represented) are not used.}
    \label{fig:SiNE}
\end{figure}

Signed Network Embedding (SiNE) is a deep learning framework for vertex embedding in signed graphs, proposed by Wang \textit{et al}.~\cite{Wang2017}. Following~\cite{Cygan2012}, it is based on the assumption that the representation of a vertex should be similar to its positive neighbors, and dissimilar to its negative ones. To model signed networks based on this principle, SiNE proceeds at a very local level by focusing on a very specific type of subgraph. It extracts the set of all \textit{open triads} (i.e., three vertices connected by two edges) present in the graph, focusing on those containing one positive and one negative edge, as illustrated by Figure~\ref{fig:SiNE}.a. This set of triads represents the graph and is fed to a deep learning framework composed of two neural networks sharing certain weights. Following the principle mentioned earlier, this model is trained to minimize the similarity between the representation of the vertex located at the center of the triad and its negative neighbors, while maximizing its similarity with its positive ones. One neural network within the framework is dedicated to the positive neighbors, while the other network handles the negative ones. 

A limitation of this approach is that it ignores vertices whose neighbors are all negative or all positive, such as the first triad in Figure~\ref{fig:SiNE}.b. To handle the latter case, SiNE introduces a dummy vertex, denoted by $u$ in the figure, that is connected through a negative edge to the central vertex of the positive triad, $v_i$. This allows creating as many dummy triads containing one positive and one negative edge, which can be processed by the framework. A similar principle could be used to handle negative triads, by adding positive dummy edges. However, Wang \textit{et al}. stress that this is not justified by the signed network literature, and prefer to discard negative triads~\cite{Wang2017}. 

At the end of the training process, SiNE learns a compact representation for each vertex in the input graph, based only on its direct neighborhood. To get a graph-level representation, we simply aggregate the representations of all the vertices in the graph. For the sake of completeness, we consider two approaches: averaging these representations and summing them. We select SiNE as a baseline because it reaches top performances on several tasks and datasets in the literature~\cite{Derr2018, Kim2018b,  Shen2020}. Of course, these are vertex-level tasks, and not graph-level tasks as in the present article. 

According to Wang \textit{et al}.~\cite{Wang2017}, the time complexity of SiNE is $O(R n S T d^2)$, where $R$ is the number of epochs, $n$ the number of vertices in the input graph, $S$ the number of triplets used to describe each vertex, $T$ the number of layers in the neural network, and $d$ the dimension of the representation (or of the layers, whichever is larger). 
In our case, we do not work with a single graph, but rather a whole collection of graphs $\mathcal{G}$. Consequently, we must include an additional multiplicative factor corresponding to the number of graphs in the collection. Moreover, we must replace $n$ by $N$, the number of vertices in the largest graph in $\mathcal{G}$. As a result, the total expression is $O(|\mathcal{G}| R N S T d^2)$. Variables $R$, $S$, $T$ and $d$ are user-controlled parameters, whereas $\mathcal{G}$ and $N$ depend on the data.

\subsection{Signed Graph2vec}
\label{sec:MethodsSG2V}
Graph2vec~\cite{Narayanan2017} is an embedding method designed to learn representations of whole \textit{unsigned} graphs, and is based on an analogy with the Doc2vec approach defined for text~\cite{Le2014}. Its principle is to consider graphs (documents, in the analogy) as collections of subgraphs (words). The procedure enumerates rooted subgraphs around all vertices of the considered graphs. Each one represents the neighborhood of a vertex (the so-called root) in a certain order. 

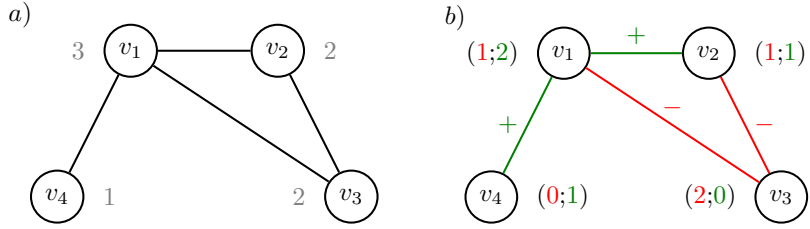
\begin{figure}[!t]
    \centering
    \hspace*{\fill}
    \resizebox{0.42\linewidth}{!}{\begin{tikzpicture}
    \tikzstyle{nn}=[circle, thick, draw=black, minimum size=7mm, fill=none]
		\node[nn] (v1) at (0, 0) {$v_1$};
		\node[nn] (v2) at (2, 0) {$v_2$};
		\node[nn] (v3) at (3, -2) {$v_3$};
		\node[nn] (v4) at (-1, -2) {$v_4$};
        
		\draw[thick] (v1) to (v2);
		\draw[thick] (v1) to (v3);
		\draw[thick] (v1) to (v4);
		\draw[thick] (v2) to (v3);

        \node[left=0.5cm, gray] at (v1) {$3$};
        \node[right=0.5cm, gray] at (v2) {$2$};
        \node[left=0.5cm, gray] at (v3) {$2$};
        \node[right=0.5cm, gray] at (v4) {$1$};

        \node at (-1.5,0.5) {$a)$};
\end{tikzpicture}}
    \hfill
    \resizebox{0.42\linewidth}{!}{\begin{tikzpicture}
    \tikzstyle{nn}=[circle, thick, draw=black, minimum size=7mm, fill=none]
		\node[nn] (v1) at (0, 0) {$v_1$};
		\node[nn] (v2) at (2, 0) {$v_2$};
		\node[nn] (v3) at (3, -2) {$v_3$};
		\node[nn] (v4) at (-1, -2) {$v_4$};
        
		\draw[thick, color=green!50!black] (v1) edge node[above] {$+$} (v2);
		\draw[thick, red] (v1) edge node[above] {$-$} (v3);
		\draw[thick, color=green!50!black] (v1) edge node[left] {$+$} (v4);
		\draw[thick, red] (v2) edge node[right] {$-$} (v3);

        \node[left=0.5cm] at (v1) {$(\textcolor{red}{1}{;}\textcolor{green!50!black}{2})$};
        \node[right=0.5cm] at (v2) {$(\textcolor{red}{1}{;}\textcolor{green!50!black}{1})$};
        \node[left=0.5cm] at (v3) {$(\textcolor{red}{2}{;}\textcolor{green!50!black}{0})$};
        \node[right=0.5cm] at (v4) {$(\textcolor{red}{0}{;}\textcolor{green!50!black}{1})$};

        \node at (-1.5,0.5) {$b)$};
\end{tikzpicture}}
    \hspace*{\fill}
    \caption{Unsigned (a) and signed (b) graphs used to illustrate the relabelling rules of G2V and SG2V (cf. text). The numeric values are the vertex degrees.}
    \label{fig:SG2V}
\end{figure}

These rooted subgraphs are named using labels obtained with the relabeling procedure of the Weisfeiler--Lehman isomorphism test~\cite{Weisfeiler1968} (WL for short). Starting with the degree as the initial vertex label, this procedure goes through two phases to iteratively update these labels. First, each vertex is described by a tuple consisting of its previous label, and a sorted multiset containing those of its neighbors. Second, each unique tuple is replaced by a new label, to be used in the next iteration. Two identical tuples are replaced by the same label, but two different ones get distinct labels. This phase allows for a compact representation of the vertices. At the end of the process, each rooted subgraph is represented by its root's label. More formally, the label update rule is
\begin{equation}
    \ell_t(u) = f\big(\ell_{t-1}(u), \{ \ell_{t-1}(v) : v \in N(u) \}\big),
\end{equation}
where $\ell_t(u)$ is the label of the subgraph rooted in $u$ at iteration $t$, $N(u)$ is the neighborhood of $u$, and $f$ is an injective function used to replace the tuples by new labels. Note that the set of the neighbors' labels is ordered. Consider $v_1$ in the left graph of Figure~\ref{fig:SG2V}, for instance. Assuming degree is used to initialize the vertex labels, the composite label produced by the above rule is \texttt{3{,}122}: the initial label of $v_1$ is \texttt{3}, and those of its neighbors are \texttt{1}, \texttt{2}, \texttt{2} when ranked in increasing order. This composite label is then fetched to $f$, which returns the compressed label. Here, it could be, for example \texttt{4}, which has not been used yet at this stage of the process. 

The number of iterations corresponds to the desired order of the neighborhood covered by the rooted subgraph (i.e., how far the subgraph spreads). At the end of this process, two isomorphic rooted subgraphs should get the same label. The obtained labels are then used to train the standard Doc2vec SkipGram model. Graph2vec has proven its effectiveness in many tasks involving the classification of \textit{unsigned} graphs~\cite{Fang2020, Zhou2022, Ngo2021}.

This method is not able to take advantage of the additional information present in signed graphs (i.e., edge signs), though. For this purpose, we define Signed Graph2vec, an adaptation of Graph2vec that relies on two variants of the WL relabeling procedure able to handle edge signs. The first one, denoted by \texttt{SG2Vn} (\texttt{n} for neutral), is straightforward and does not assume that the network has any form of structural balance. Regarding label initialization, instead of using the degree, we use both positive and negative degrees. For each vertex $u$, we first define a tuple $\big(k^+(u);k^-(u)\big)$, and then replace each unique pair with a unique label using $f$. For the rest of the iterations, we proceed as in Graph2vec, except that we append the sign of the concerned edge in front of each neighbor when building the labels:
\begin{equation}
    \ell_t(u) = f \Big( \ell_{t-1}(u), \big\{ \big[ s(u,v), \ell_{t-1}(v) \big] : v \in N(u) \big\} \Big),
\end{equation}
where $s(u,v)$ is the sign of edge $(u,v)$, and $[~]$ denotes string concatenation. This allows distinguishing vertices holding the same label but connected to the root with edges of opposed signs. In our example from Figure~\ref{fig:SG2V} (right), applying $f$ to the pairs of negative and positive degrees could produce the initial labels \texttt{1}, \texttt{2}, \texttt{3} and \texttt{4} (in this order), as all these pairs are different. According to the above rule, the composite label of $v_1$ at the first iteration would be \texttt{1{,}+2-3+4}. 

The second relabeling method, denoted by \texttt{SG2Vsb}, is the one proposed (for a different purpose) by Zhang \textit{et al}. in~\cite{Zhang2023}, and it assumes that the network is structurally balanced. Each vertex is represented by two labels, based on its positive and negative reachable sets, respectively (cf. Section~\ref{sec:BackSignedGraphs}). The authors do not explain how they perform their initialization, so we use the positive (resp. negative) degree for the positive (resp. negative) label. The method requires one update rule for each label:
\begin{align}
    \ell_t^+(u) &= f\big(\ell_{t-1}^+(u), \{ \ell_{t-1}^+(v) : v \in N^+(u) \}, \{ \ell_{t-1}^-(v) : v \in N^-(u) \}\big) \\
    \ell_t^-(u) &= f\big(\ell_{t-1}^-(u), \{ \ell_{t-1}^-(v) : v \in N^+(u) \}, \{ \ell_{t-1}^+(v) : v \in N^-(u) \}\big).
\end{align}
In our example from Figure~\ref{fig:SG2V} (right), the positive composite label of $v_1$ at the first iteration is \texttt{2{,}11{,}2}, because it has an initial positive label of \texttt{2}, two positive neighbors with an initial positive label of \texttt{1}, and one negative neighbor with an initial negative label of \texttt{2}. Symmetrically, the negative label of $v_1$ is \texttt{1{,}01{,}0}. 
At the end of the process, $f$ is applied to tuples formed by the positive and negative labels of each vertex, resulting in the final rooted subgraph labels.

Based on the algorithm described in~\cite{Narayanan2017}, the time complexity of Graph2Vec is $O(|\mathcal{G}| R N T^2 d)$, where $|\mathcal{G}|$ is the size of the collection, $R$ the number of epochs, $N$ the number of vertices of the largest graph in $\mathcal{G}$, $T$ the order of the neighborhood covered by the rooted subgraphs, and $d$ the dimension of the representation. The complexity is identical for our signed generalization, since our modifications do not affect the computational cost of the original algorithm. Like for SiNE, time complexity is linear in the collection size $\mathcal{G}$ and in the graph order $N$, but it also depends on some user-controlled parameters ($R$, $T$, and $d$). 


\subsection{Whole-graph Signed GCN}
\label{sec:MethodsSGCN}
Graph Convolutional Networks (GCNs) are a family of neural networks that adapt traditional Convolutional Neural Networks (CNNs) so that they can process graph data instead of standard tabular data~\cite{Wu2020f}. Shallow embedding methods such as Graph2vec may miss complex patterns in the graphs, and deep learning methods like CNNs are likely to solve this limitation~\cite{Zhang2018ak}. GCNs generalize the convolution operation by considering graph neighborhoods instead of linear or grid neighborhoods as in standard CNNs used in NLP and image processing. 

Most GCNs are designed to produce \textit{vertex}-level representations. The general principle is as follows. Each vertex is initially represented by a vector, that can be generated randomly or based on some vertex features. In a convolution layer, the representation of a given vertex is combined with that of its neighbors. The result is then fed to a generally non-linear function (e.g., multilayer perceptron) to get the updated vertex representation. The information is propagated through multiple layers to incorporate information from multi-hop neighbors. The maximal number of hops corresponds to the number of convolution layers in the network. GCNs achieve state-of-the-art performances on many tasks~\cite{Derr2018} such as vertex classification, edge prediction, and community detection.

Standard GCNs are only able to handle \textit{unsigned} networks, though. Leveraging the information conveyed by edge signs mainly requires adapting the message-passing rules used when computing the graph-based convolution. A few methods allow doing so, and in this work, we focus on \textit{Signed Graph Convolutional Networks} (SGCN)~\cite{Derr2018}, for two reasons. First, SGCN obtains strong performances on many tasks, such as node classification and link sign prediction, and is used as a basis for multiple other methods~\cite{Tang2022, Kim2023, Ma2021}. Second, its implementation is conveniently available online\footnote{\url{https://github.com/benedekrozemberczki/SGCN}} and can be modified to fit our needs. 
Similarly to \texttt{SG2Vsb} (Section~\ref{sec:MethodsSG2V}), this method relies on a dual hidden representation of a vertex, corresponding to its positive vs. negative reach sets, and it uses balance theory to aggregate and propagate vertex representation across layers. Formally, the hidden representations $\mathbf{h}$ are updated as follows:
\begin{align}
    \mathbf{h}_t^+(u) &= \sigma\bigg( \mathbf{W}_t^+ \Big[ \sum_{v \in N^+} \frac{\mathbf{h}_{t-1}^+(v)}{k^+(u)}, \sum_{v \in N^-} \frac{\mathbf{h}_{t-1}^-(v)}{k^-(u)}, \mathbf{h}^+_{t-1}(u) \Big] \bigg) \\
    \mathbf{h}_t^-(u) &= \sigma\bigg( \mathbf{W}_t^- \Big[ \sum_{v \in N^+} \frac{\mathbf{h}_{t-1}^-(v)}{k^+(u)}, \sum_{v \in N^-} \frac{\mathbf{h}_{t-1}^+(v)}{k^-(u)}, \mathbf{h}^-_{t-1}(u) \Big] \bigg),
\end{align}
where $\sigma$ is a non-linear activation function, $[~]$ denotes the concatenation, and the $\mathbf{W}$ matrices are learnable weights. Ultimately, the dual hidden representations are concatenated to obtain a single vertex representation.

\begin{figure}[!t]
    \centering
    \resizebox{1\linewidth}{!}{\input{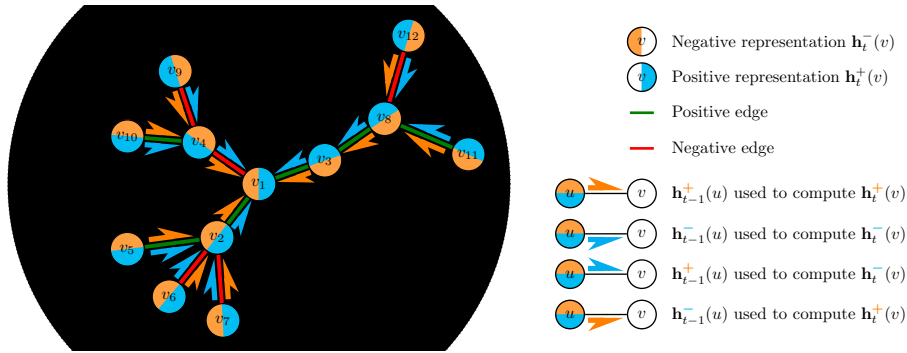}}
    \caption{Update rules of SGCN, applied to an example graph. Only the edges belonging to a shortest path between the vertex of interest $v_1$ and some other vertex are shown.}
    \label{fig:SGCN}
\end{figure}

Figure~\ref{fig:SGCN} illustrates how these update rules work, focusing on vertex $v_1$. The dual representation of the vertices is shown using two colors: orange (negative) and cyan (positive). The update rules are depicted by pairs of arrows. Each arrow is attached either to the positive or negative representation of the \textit{source} vertex, and shows how it is used to update the representation of the target vertex. The arrow color indicates if this update affects the negative (orange) or positive (cyan) representation of the \textit{target} vertex (see the figure legend). 
Coming back to $v_1$, the first convolution layer combines the positive representations of its positive neighbors $v_2$ and $v_3$, as well as the negative representation of its negative neighbor $v_4$, to build $\mathbf{h}_1^+(v_1)$. Symmetrically, $\mathbf{h}_1^-(v_1)$ is based on the negative representations of $v_2$ and $v_3$ and the positive representation of $v_4$. 
The second layer performs an update that takes into account the representations of the second-order neighbors, in a way that respects structural balance. For instance, since $v_9$ is connected to $v_1$ by a positive path, its positive representation is (indirectly) used to compute $\mathbf{h}_2^+(v_1)$, and its negative representation to compute $\mathbf{h}_2^-(v_1)$. On the contrary, $v_{10}$ is connected to $v_1$ by a negative path, so its positive and negative representations affect $\mathbf{h}_2^-(v_1)$ and $\mathbf{h}_2^+(v_1)$, respectively. 

SGCN effectively learns representations of vertices in signed graphs. However, our objective is to handle \textit{whole} graphs. A solution proposed for unsigned graphs in the literature~\cite{Gilmer2017} consists of adding a \textit{master node} (also called a \textit{virtual} or \textit{super} node), which is connected to all other vertices in the graph. One then uses the representation of this master node as the representation of the whole graph. The intuition is that, as the master node is connected to all parts of the graph, its representation aggregates all its information. 

\begin{figure}[!t]
    \centering
    \resizebox{0.32\linewidth}{!}{\begin{tikzpicture}

    \node [nn] (v1) at (-0.50,  1.50) {$v_1$};
    \node [nn] (v2) at ( 0.00,  0.00) {$v_2$};
    \node [nn] (v3) at ( 1.50, -1.00) {$v_3$};
    \node [nn] (v4) at ( 3.00, -1.00) {$v_4$};
    \node [nn] (v5) at ( 4.50,  0.00) {$v_5$};
    \node [nn] (v6) at ( 5.00,  1.50) {$v_6$};
    \node [rr,fill=green!50!black!20] (MN+) at (2.25, 3.00) {$MN_+$};

    \draw[color=green!50!black] (v1) edge (v2);
    \draw[color=red] (v1) edge (v3);
    \draw[color=red] (v2) edge (v3);
    \draw[color=red] (v2) edge (v3);
    \draw[color=green!50!black] (v2) edge (v4);
    \draw[color=green!50!black] (v3) edge (v4);
    \draw[color=red] (v4) edge (v5);
    \draw[color=red] (v4) edge (v6);
    \draw[color=green!50!black] (v5) edge (v6);

    \draw[dashed, color=green!50!black] (MN+) edge (v1);
    \draw[dashed, color=green!50!black] (MN+) edge (v2);
    \draw[dashed, color=green!50!black] (MN+) edge (v3);
    \draw[dashed, color=green!50!black] (MN+) edge (v4);
    \draw[dashed, color=green!50!black] (MN+) edge (v5);
    \draw[dashed, color=green!50!black] (MN+) edge (v6);
    
    \node at (-0.50,3.00) {\fontsize{15}{15}\selectfont$a)$};
\end{tikzpicture}}
    \hfill
    \resizebox{0.32\linewidth}{!}{\begin{tikzpicture}

    \node [nn] (v1) at (-0.50,  1.50) {$v_1$};
    \node [nn] (v2) at ( 0.00,  0.00) {$v_2$};
    \node [nn] (v3) at ( 1.50, -1.00) {$v_3$};
    \node [nn] (v4) at ( 3.00, -1.00) {$v_4$};
    \node [nn] (v5) at ( 4.50,  0.00) {$v_5$};
    \node [nn] (v6) at ( 5.00,  1.50) {$v_6$};
    \node [rr,fill=red!20] (MN-) at (2.25, 3.00) {$MN_-$};

    \draw[color=green!50!black] (v1) edge (v2);
    \draw[color=red] (v1) edge (v3);
    \draw[color=red] (v2) edge (v3);
    \draw[color=red] (v2) edge (v3);
    \draw[color=green!50!black] (v2) edge (v4);
    \draw[color=green!50!black] (v3) edge (v4);
    \draw[color=red] (v4) edge (v5);
    \draw[color=red] (v4) edge (v6);
    \draw[color=green!50!black] (v5) edge (v6);

    \draw[dashed, color=red] (MN-) edge (v1);
    \draw[dashed, color=red] (MN-) edge (v2);
    \draw[dashed, color=red] (MN-) edge (v3);
    \draw[dashed, color=red] (MN-) edge (v4);
    \draw[dashed, color=red] (MN-) edge (v5);
    \draw[dashed, color=red] (MN-) edge (v6);
    
    \node at (-0.50,3.00) {\fontsize{15}{15}\selectfont$b)$};
\end{tikzpicture}}
    \hfill
    \resizebox{0.32\linewidth}{!}{\begin{tikzpicture}

    \node [nn] (v1) at (-0.50,  1.50) {$v_1$};
    \node [nn] (v2) at ( 0.00,  0.00) {$v_2$};
    \node [nn] (v3) at ( 1.50, -1.00) {$v_3$};
    \node [nn] (v4) at ( 3.00, -1.00) {$v_4$};
    \node [nn] (v5) at ( 4.50,  0.00) {$v_5$};
    \node [nn] (v6) at ( 5.00,  1.50) {$v_6$};
    \node [rr,fill=green!50!black!20] (MN+) at (1.25, 3.00) {$MN_+$};
    \node [rr,fill=red!20] (MN-) at (3.25, 3.00) {$MN_-$};

    \draw[color=green!50!black] (v1) edge (v2);
    \draw[color=red] (v1) edge (v3);
    \draw[color=red] (v2) edge (v3);
    \draw[color=red] (v2) edge (v3);
    \draw[color=green!50!black] (v2) edge (v4);
    \draw[color=green!50!black] (v3) edge (v4);
    \draw[color=red] (v4) edge (v5);
    \draw[color=red] (v4) edge (v6);
    \draw[color=green!50!black] (v5) edge (v6);

    \draw[dashed, color=green!50!black] (MN+) edge (v1);
    \draw[dashed, color=green!50!black] (MN+) edge (v2);
    \draw[dashed, color=green!50!black] (MN+) edge (v3);
    \draw[dashed, color=green!50!black] (MN+) edge (v4);
    \draw[dashed, color=green!50!black] (MN+) edge (v5);
    \draw[dashed, color=green!50!black] (MN+) edge (v6);

    \draw[dashed, color=red] (MN-) edge (v1);
    \draw[dashed, color=red] (MN-) edge (v2);
    \draw[dashed, color=red] (MN-) edge (v3);
    \draw[dashed, color=red] (MN-) edge (v4);
    \draw[dashed, color=red] (MN-) edge (v5);
    \draw[dashed, color=red] (MN-) edge (v6);
    
    \node at (-0.50,3.00) {\fontsize{15}{15}\selectfont$c)$};
\end{tikzpicture}}
    \\[0.5cm]
    \resizebox{0.40\linewidth}{!}{\begin{tikzpicture}
    
    \fill[fill=blue!20,rotate around={-35:(1.10,0.10)}] (1.10,0.10) ellipse (3.0cm and 1.5cm);
    \fill[fill=orange!20,rotate around={ 75:(4.75,0.75)}] (4.75,0.75) ellipse (1.5cm and 0.9cm);
    
    \node [nn] (v1) at (-0.50,  1.50) {$v_1$};
    \node [nn] (v2) at ( 0.00,  0.00) {$v_2$};
    \node [nn] (v3) at ( 1.50, -1.00) {$v_3$};
    \node [nn] (v4) at ( 3.00, -1.00) {$v_4$};
    \node [nn] (v5) at ( 4.50,  0.00) {$v_5$};
    \node [nn] (v6) at ( 5.00,  1.50) {$v_6$};
    \node [rr,fill=blue!20] (MN1) at (1.25, 3.00) {$MN_1$};
    \node [rr,fill=orange!20] (MN2) at (3.25, 3.00) {$MN_2$};

    \draw[color=green!50!black] (v1) edge (v2);
    \draw[color=red] (v1) edge (v3);
    \draw[color=red] (v2) edge (v3);
    \draw[color=red] (v2) edge (v3);
    \draw[color=green!50!black] (v2) edge (v4);
    \draw[color=green!50!black] (v3) edge (v4);
    \draw[color=red] (v4) edge (v5);
    \draw[color=red] (v4) edge (v6);
    \draw[color=green!50!black] (v5) edge (v6);

    \draw[dashed, very thin, color=green!50!black] (MN1) edge (v1);
    \draw[dashed, very thin, color=green!50!black] (MN1) edge (v2);
    \draw[dashed, very thin, color=green!50!black] (MN1) edge (v3);
    \draw[dashed, very thin, color=green!50!black] (MN1) edge (v4);
    \draw[dashed, very thin, color=red] (MN1) edge (v5);
    \draw[dashed, very thin, color=red] (MN1) edge (v6);
    
    \draw[dashed, very thin, color=red] (MN2) edge (v1);
    \draw[dashed, very thin, color=red] (MN2) edge (v2);
    \draw[dashed, very thin, color=red] (MN2) edge (v3);
    \draw[dashed, very thin, color=red] (MN2) edge (v4);
    \draw[dashed, very thin, color=green!50!black] (MN2) edge (v5);
    \draw[dashed, very thin, color=green!50!black] (MN2) edge (v6);
    
    \node at (-0.50,3.00) {\fontsize{15}{15}\selectfont$d)$};
\end{tikzpicture}}
    \hfill
    \resizebox{0.40\linewidth}{!}{\begin{tikzpicture}
    
    \fill[fill=blue!20,rotate around={-75:(-0.25,0.75)}] (-0.25,0.75) ellipse (1.4cm and 0.9cm);
    \fill[fill=orange!20,rotate around={00:(2.25,-1.00)}] (2.25,-1.00) ellipse (1.5cm and 0.9cm);
    \fill[fill=purple!20,rotate around={75:(4.75,0.75)}] (4.75,0.75) ellipse (1.5cm and 0.9cm);
    
    \node [nn] (v1) at (-0.50,  1.50) {$v_1$};
    \node [nn] (v2) at ( 0.00,  0.00) {$v_2$};
    \node [nn] (v3) at ( 1.50, -1.00) {$v_3$};
    \node [nn] (v4) at ( 3.00, -1.00) {$v_4$};
    \node [nn] (v5) at ( 4.50,  0.00) {$v_5$};
    \node [nn] (v6) at ( 5.00,  1.50) {$v_6$};
    \node [rr,fill=blue!20] (MN1) at (0.75, 3.00) {$MN_1$};
    \node [rr,fill=orange!20] (MN2) at (2.25, 3.00) {$MN_2$};
    \node [rr,fill=purple!20] (MN3) at (3.75, 3.00) {$MN_3$};

    \draw[color=green!50!black] (v1) edge (v2);
    \draw[color=red] (v1) edge (v3);
    \draw[color=red] (v2) edge (v3);
    \draw[color=red] (v2) edge (v3);
    \draw[color=green!50!black] (v2) edge (v4);
    \draw[color=green!50!black] (v3) edge (v4);
    \draw[color=red] (v4) edge (v5);
    \draw[color=red] (v4) edge (v6);
    \draw[color=green!50!black] (v5) edge (v6);

    \draw[dashed, very thin, color=green!50!black] (MN1) edge (v1);
    \draw[dashed, very thin, color=green!50!black] (MN1) edge (v2);
    \draw[dashed, very thin, color=red] (MN1) edge (v3);
    \draw[dashed, very thin, color=red] (MN1) edge (v4);
    \draw[dashed, very thin, color=red] (MN1) edge (v5);
    \draw[dashed, very thin, color=red] (MN1) edge (v6);
    
    \draw[dashed, very thin, color=red] (MN2) edge (v1);
    \draw[dashed, very thin, color=red] (MN2) edge (v2);
    \draw[dashed, very thin, color=green!50!black] (MN2) edge (v3);
    \draw[dashed, very thin, color=green!50!black] (MN2) edge (v4);
    \draw[dashed, very thin, color=red] (MN2) edge (v5);
    \draw[dashed, very thin, color=red] (MN2) edge (v6);
    
    \draw[dashed, very thin, color=red] (MN3) edge (v1);
    \draw[dashed, very thin, color=red] (MN3) edge (v2);
    \draw[dashed, very thin, color=red] (MN3) edge (v3);
    \draw[dashed, very thin, color=red] (MN3) edge (v4);
    \draw[dashed, very thin, color=green!50!black] (MN3) edge (v5);
    \draw[dashed, very thin, color=green!50!black] (MN3) edge (v6);
    
    \node at (-0.50,3.00) {\fontsize{15}{15}\selectfont$e)$};
\end{tikzpicture}}
    \hfill
    \resizebox{0.18\linewidth}{!}{\begin{tikzpicture}
    
    \node [nn, anchor=east] at (0.00, 0.80) {$v$};
    \node [anchor=west] at (0.25, 0.80) {Graph vertex};
    
    \draw[color=green!50!black, ultra thick] (-0.50,0.00) edge (0.00,0.00);
    \node [anchor=west] at (0.25,0.00) {Positive graph edge};
    
    \draw[color=red, ultra thick] (-0.50,-0.80) edge (0.00,-0.80);
    \node [anchor=west] at (0.25,-0.80) {Negative graph edge};
    
    \node [rr,fill=black!20,anchor=east] at (0.00,-1.60) {$MN$};
    \node [anchor=west] at (0.25,-1.60) {Master node};
    
    \draw[dashed, color=green!50!black, ultra thick] (-0.50,-2.40) edge (0.00,-2.40);
    \node [anchor=west] at (0.25, -2.40) {Positive master node link};
    
    \draw[dashed, color=red, ultra thick] (-0.50,-3.20) edge (0.00,-3.20);
    \node [anchor=west] at (0.25, -3.20) {Negative master node link};
    
    \fill[fill=black!20,rotate around={55:(-0.25,-4.00)}] (-0.25,-4.00) ellipse (0.6cm and 0.3cm);
    \node [anchor=west] at (0.25, -4.00) {Cluster};
    
    \node at (0.00,-6.00) {\textcolor{white}{x}};
\end{tikzpicture}}
    
    \caption{Examples of the 5 proposed interconnection schemes: \texttt{WSGCN+} (a), \texttt{WSGCN-} (b), \texttt{WSGCN±} (c), \texttt{WSGCNsb} (d) and \texttt{WSGCNgb} (e). \textit{MN} stands for Master Node. Green and red edges represent positive and negative connections, respectively. Each colored ellipse is a cluster.}
    \label{fig:SGCNConnectionScheme}
\end{figure}
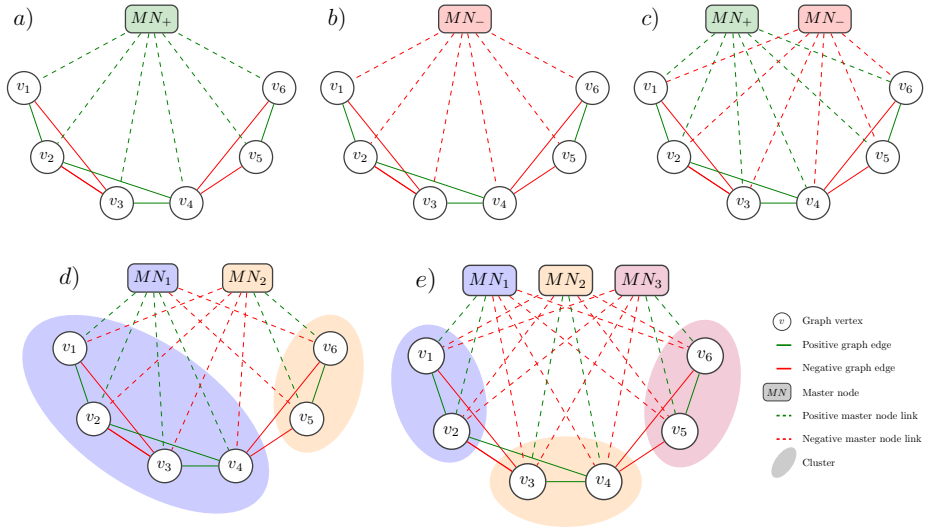

It is not straightforward to adapt this approach to signed graphs, though: connecting a master node to the rest of the graph is not trivial, as there are two different types of edges. We propose five interconnection schemes, which we call collectively \textit{Whole-graph SGCN} (WSGCN). They are represented in Fig.~\ref{fig:SGCNConnectionScheme}: circular vertices and solid lines correspond to the original graph, whereas the master nodes are shown as rounded rectangles denoted by $MN$, and their connections to the original vertices are shown as dotted lines. The first three interconnection schemes do not respect structural balance and can be considered baselines: \texttt{WSGCN+} (Fig.~\ref{fig:SGCNConnectionScheme}.a) and \texttt{WSGCN-} (Fig.~\ref{fig:SGCNConnectionScheme}.b) consist of connecting a single master node to the rest of the graph using only positive and negative edges, respectively. With \texttt{WSGCN±} (Fig.~\ref{fig:SGCNConnectionScheme}.c), we use two distinct master nodes (one positive and one negative), which allows us to combine both previous schemes at once. The whole-graph representation is obtained by summing both master node representations. The fourth scheme, \texttt{WSGCNsb} (Fig.~\ref{fig:SGCNConnectionScheme}.d), is based on a (strict) structural balance. Using the \textit{signnet} library\footnote{\url{https://github.com/schochastics/signnet}}, we detect the optimal graph bisection. Then, we connect one distinct master node to each cluster by enforcing structural balance, i.e., positive edges within the cluster and negative ones within the other cluster. Like before, the whole-graph embedding is obtained by summing the representations of both master nodes. Finally, the fifth scheme, \texttt{WSGCNgb} (Fig.~\ref{fig:SGCNConnectionScheme}.e), relies on the \textit{generalized} structural balance. As before, we find the optimal graph partition, but this time there may be more than two clusters. We add one master node for each cluster and sum their representations to get the whole-graph embedding.

SGCN generalizes a vanilla GCN architecture by using a dual representation of the vertices. This does not affect the time complexity of such architecture, which is $O(R T (m d + n d^2))$ according to the literature~\cite{Wu2020f, Chen2020s}, where $R$ is the number of epochs, $T$ the number of layers in the neural network, $m$ the size of the graph (i.e., its number of edges), $n$ its order (i.e., number of vertices), and $d$ the dimension of the representation. The master nodes that we introduce in our whole-graph adaptation of SGCN do not imply any change in this complexity either. However, we iteratively apply our WSGCN method to all graphs in a collection $|\mathcal{G}|$, so its total time complexity is $O(|\mathcal{G}| R T (M d + N d^2))$, with $M$ the size of the largest graph, and $N$ its order. Like for the other methods described in this section, some of these variables correspond to user-controlled parameters ($R$, $T$, and $d$), whereas others depend on the data ($\mathcal{G}$, $M$, and $N$).


\section{Results}
\label{sec:Results}
Our experimental protocol consists of assessing all the methods described in Section~\ref{sec:Methods} on the datasets and tasks presented in Section~\ref{sec:Datasets}. We produce the representations, then train an SVM to perform the classification, using a $10$-fold cross-validation. Our code is available online\footnote{\url{https://github.com/CompNet/SWGE} \label{ftn:Software}}. We conduct our experiments on an Nvidia RTX 2080 Ti GPU.

We discuss each family of methods separately: SiNE (Section~\ref{sec:ResEmbeddings}), Graph2vec (Section~\ref{sec:ResG2V}), and Signed Graph Convolutional Networks (Section~\ref{sec:ResSGCN}); before comparing them (Section~\ref{sec:ResComp}). All the performance values are expressed in terms of the macro-averaged $F$-measure, i.e., by computing the $F$-measure for each class separately, then averaging them to get the overall performance. This allows giving the same importance to all classes, even in imbalanced datasets. For the sake of completeness, performance expressed in terms of Precision and Recall is provided in the Appendix.

\subsection{Aggregated Signed Network Embedding}
\label{sec:ResEmbeddings}
In this section, we present the results obtained with SiNE, the signed vertex embedding method that we consider our baseline. The performance scores are shown in Table~\ref{tab:ResultsSiNE} for all our datasets, in terms of macro $F$-measure. Each column focuses on one of the two functions used to aggregate vertex representations: sum and average. The corresponding tables describing performance in terms of Precision (Table~\ref{tab:ResultsPreSiNE}) and Recall (Table~\ref{tab:ResultsRecSiNE}) are available in the Appendix. Overall, the Precision tends to be slightly higher than the Recall for the SSO and CCS datasets, whereas the opposite is true for the EPF dataset. However, both metrics follow the trends described below for the $F$-measure. 

\begin{table}[!t]
    \caption{Results in terms of macro $F$-measure obtained with \texttt{SiNE}. Each row focuses on a task, whereas columns represent the function used to aggregate vertex representations.}
    \label{tab:ResultsSiNE}
    \centering
    \begin{tabularx}{\linewidth}{X r r}
        \hline
        \textbf{Task} & \textbf{Sum} & \textbf{Average} \\
        \hline
        SSO & $\mathbf{55.42}$ & $50.21$ \\
        CCS & $\mathbf{50.48}$ & $48.62$ \\
        EPF & $\mathbf{69.81}$ & $67.58$ \\
        \hline
    \end{tabularx}
\end{table}

On SSO and CCS, the performances are close to a random classifier, which would get an expected $50\%$ $F$-measure. SiNE performs better on EPF, with the best result reaching $69.81\%$. We assume that for this task, the local information available is often sufficiently discriminative. Moreover, as the graphs in this dataset tend to be larger, SiNE can extract a greater number of training triads, contributing to an improved model performance. Furthermore, the choice of the aggregation function has a notable impact on performance, with the sum operation yielding better results across all tasks.

\subsection{Signed Graph2vec}
\label{sec:ResG2V}
In this section, we present the results obtained with our two proposed variants of the \textit{Signed Graph2vec} method: \texttt{SG2Vn} (which does not enforce SB) and \texttt{SG2Vsb} (which does). As a reference, we also include in our study Graph2vec (noted \texttt{G2V}), which simply ignores all edge signs. The performances are shown in Table~\ref{tab:ResultsG2V}, in terms of macro $F$-measure, whereas scores expressed in terms of Precision (Table~\ref{tab:ResultsPreG2V}) and Recall (Table~\ref{tab:ResultsRecG2V}) are provided in the Appendix. The columns show the performance for an increasing number of iterations, ranging from 1 to 5. As a reminder, this parameter controls the order of the rooted subgraphs extracted to describe the graph and, therefore, the range that is taken into account when characterizing vertex neighborhoods. The Precision and the Recall scores are very comparable, which is why we now focus on the $F$-measure. In the following, we discuss each dataset separately. 

\subsubsection{Signed Space Origin}
The top part of Table~\ref{tab:ResultsG2V} presents the performances of the three Graph2vec variants on the \textit{Signed SpaceOrigin} dataset. It appears that they are affected diversely by the number of iterations. On the one hand, \texttt{G2V} and \texttt{SG2Vn} get their best score with a single iteration, and increasing iterations tend to reduce the performance. On the other hand, \texttt{SG2Vsb} starts low but increases consistently with the number of iterations. In the end, it reaches an $F$-measure of $77.44\%$ and outperforms both other variants. The average diameter is $5.47$ in this dataset, which means five iterations correspond to almost complete graph coverage.

From these observations, we can assume that a lot of information is already conveyed by the unsigned graph structures, in this dataset. This is consistent with the results obtained in the article that published the original data, as its authors already had some success performing a similar task with the unsigned version of the graphs~\cite{Papegnies2019}. Nevertheless, the signs bring some additional discriminative power, which \texttt{SG2Vsb} leverages to improve the classification performance. By comparison, this is not the case with \texttt{SG2Vn}. The relatively low level of SB Frustration in this dataset ($0.30$) may favor methods relying on SB.

\begin{table}[!t]
    \caption{Results in terms of macro $F$-measure obtained with \texttt{Graph2vec} and our two proposed signed adaptations. Each column focuses on a specific number of iterations used when extracting rooted subgraphs.}
    \label{tab:ResultsG2V}
    
    \centering
    \begin{tabularx}{\linewidth}{X l r r r r r r r}
        \hline
        \textbf{Task} & \textbf{Method} & \textbf{1 it.} & \textbf{2 it.} & \textbf{3 it.} & \textbf{4 it.} & \textbf{5 it.} \\
        \hline
        SSO & \texttt{G2V} & $75.09$ & $71.32$ & $72.62$ & $73.96$ & $73.77$ \\
        & \texttt{SG2Vn} & $74.85$ & $71.44$ & $72.15$ & $72.88$ & $72.37$ \\
        & \texttt{SG2Vsb} & $67.29$ & $72.01$ & $74.88$ & $76.98$ & $\mathbf{77.44}$ \\
        \hline
        CCS & \texttt{G2V} & $48.43$ & $48.12$ & $45.97$ & $51.85$ & $52.07$ \\
        & \texttt{SG2Vn} & $49.70$ & $49.25$ & $48.37$ & $\mathbf{52.57}$ & $52.12$ \\
        & \texttt{SG2Vsb} & $49.84$ & $51.81$ & $49.70$ & $51.37$ & $51.62$ \\
        \hline
        EPF & \texttt{G2V} & $45.63$ & $49.76$ & $52.31$ & $60.43$ & $63.14$ \\
        & \texttt{SG2Vn} & $75.63$ & $81.44$ & $84.18$ & $86.16$ & $88.68$ \\
        & \texttt{SG2Vsb} & $80.83$ & $82.90$ & $86.31$ & $87.99$ & $\mathbf{89.98}$ \\
        \hline
    \end{tabularx}
\end{table}

\subsubsection{Correlation Clustering Solutions}
The results obtained for this dataset are shown in the middle part of Table~\ref{tab:ResultsG2V}. The performance is clearly lower, for all three variants, and even similar to the score expected from a random classifier ($50$), with a maximum $F$-measure of $52.57\%$ obtained by \texttt{SG2Vn}. As explained in Section~\ref{sec:DataCorrClust}, the dataset contains two subsets: some graphs are completely connected, and the rest are not. Assuming that one type of graph might be more difficult to handle than the other, we try training separately on these subsets. However, we do not see any significant difference between the obtained results, which are similar to those already presented in Table~\ref{tab:ResultsG2V}.

We assume that either this classification task is too hard, in the sense that the information available in the graphs is not sufficient to perform the prediction, or that none of the three Graph2vec variants manage to capture the relevant information. The value predicted in this task is directly related to the distribution of edge signs (cf. Section~\ref{sec:DataCorrClust}), so we know with certainty that the information conveyed by signs is essential. The fact that \texttt{G2V} has similar performance to its signed counterparts hints at the second assumption (methods are unable to capture relevant information). In addition, the higher level of SB Frustration ($0.37$) may hinder the performance of \texttt{SG2Vsb}, compared to \texttt{SG2Vn}.

Increasing the number of iterations eventually improves the performance, but the effect is not as strong and stable as for the previous dataset. The graphs are more compact in this dataset, with an average diameter of only $3.6$, which may partially explain this observation. Indeed, a few iterations are enough to retrieve all available information, and increasing their number does not bring any new neighbors.

\subsubsection{European Parliament Factions}
The results for this dataset are shown in the bottom part of Table~\ref{tab:ResultsG2V}. It appears that both our signed adaptations perform drastically better than the original unsigned method. The difference in $F$-measure is the largest of the three datasets: $63.14$ (\texttt{G2V}) vs. $88.68$ (\texttt{SG2Vn}) and $89.98$ (\texttt{SG2Vsb}). Edge signs thus appear to be even more important for this task than they were for the \textit{Signed SpaceOrigin} dataset. The SB-based variant \texttt{SG2Vsb} is slightly above \texttt{SG2Vn}, which seems to indicate that this type of structure may be relevant to this classification task. The average SB Frustration is $0.28$ for this dataset.

This dataset contains larger graphs than the others, with an average of $67.34$ vertices (vs. $47.74$ and $27.31$ previously) and a mean diameter of $4.12$. This may explain the very strong effect of the number of iterations on the performance, even for \texttt{G2V}, the unsigned variant. It seems that even a small increase (proportionally to the order of the network) in the part of the graph covered when extracting rooted subgraphs, is enough to greatly improve the quality of the classification.

\subsection{Whole-graph Signed GCN}
\label{sec:ResSGCN}
In this section, we present the results obtained with the five variants that we proposed for the WSGCN method. Each one relies on the addition of one or several master nodes, through different interconnection schemes. The first three (\texttt{WSGCN+}, \texttt{WSGCN-}, \texttt{WSGCN±}) ignore any type of structural balance, whereas the others enforce, respectively, SB (\texttt{WSGCNsb}) and GB (\texttt{WSGCNgb}). We considered two methods to extract a graph representation: using only the last layer vs. the sum of all layers. Preliminary experiments showed that the former performs better, so we only focus on this approach in the following discussion.

We also include \texttt{SGCN} as a reference in our study, i.e., the original method without any master node. To get a graph-level representation, we proceed like for SiNE, and sum the representations of all vertices. We alternatively experimented with averaging them, but got lower performance, which is why we focus only on the sum, here. The results are shown in Table~\ref{tab:ResultsSGCN}, in terms of macro $F$-measure. As before, the performance in terms of Precision (Table~\ref{tab:ResultsPreSGCN}) and Recall (Table~\ref{tab:ResultsRecSGCN}) is provided in the Appendix. The columns show the results for an increasing number of convolution layers, ranging from 1 to 5. When comparing Precision and Recall, we observe very similar scores, as with the previous methods, which is why we focus on the $F$-measure results. In the following, we discuss each dataset separately. 

\begin{table}[!t]
    \caption{Results in terms of macro $F$-measure obtained with the original \texttt{SGCN} and our five proposed \texttt{WSGCN} interconnection schemes. Each column focuses on a specific number of convolution layers.}
    \label{tab:ResultsSGCN}
    
    \centering
    \begin{tabularx}{\linewidth}{X l r r r r r}
        \hline
        \textbf{Task} & \textbf{Method} & \textbf{1 lay.} & \textbf{2 lay.} & \textbf{3 lay.} & \textbf{4 lay.} & \textbf{5 lay.} \\
        \hline
        SSO & \texttt{SGCN} & $66.48$ & $67.21$ & $68.06$ & $68.87$ & $69.54$  \\
        & \texttt{WSGCN+} & $65.12$ & $66.78$ & $68.42$ & $68.98$ & $69.29$  \\
        & \texttt{WSGCN-} & $54.19$ & $54.89$ & $55.56$ & $55.59$ & $55.89$  \\
        & \texttt{WSGCN±} & $49.28$ & $48.95$ & $49.01$ & $49.25$ & $49.39$  \\
        & \texttt{WSGCNsb} & $52.69$ & $54.99$ & $55.49$ & $55.08$ & $55.67$  \\
        & \texttt{WSGCNgb} & $66.59$ & $68.85$ & $71.21$ & $72.28$ & $\mathbf{73.69}$  \\
        \hline
        CCS & \texttt{SGCN} & $70.29$ & $70.65$ & $71.27$ & $71.43$ & $71.89$ \\
        & \texttt{WSGCN+} & $70.50$ & $70.86$ & $71.13$ & $71.27$ & $71.35$ \\
        & \texttt{WSGCN-} & $70.14$ & $70.55$ & $70.84$ & $71.02$ & $71.18$ \\
        & \texttt{WSGCN±} & $70.76$ & $71.02$ & $71.00$ & $71.20$ & $71.37$ \\
        & \texttt{WSGCNsb} & $69.59$ & $70.12$ & $70.86$ & $71.21$ & $71.46$ \\
        & \texttt{WSGCNgb} & $71.75$ & $72.20$ & $72.98$ & $73.24$ & $\mathbf{73.49}$ \\
        \hline
        EPF & \texttt{SGCN} & $90.16$ & $90.87$ & $91.63$ & $92.04$ & $92.65$ \\
        & \texttt{WSGCN+} & $88.56$ & $89.30$ & $90.49$ & $91.11$ & $91.87$ \\
        & \texttt{WSGCN-} & $88.61$ & $89.30$ & $90.49$ & $91.09$ & $91.80$ \\
        & \texttt{WSGCN±} & $86.32$ & $88.45$ & $90.01$ & $90.78$ & $91.56$ \\
        & \texttt{WSGCNsb} & $91.11$ & $92.09$ & $93.31$ & $94.17$ & $94.99$ \\
        & \texttt{WSGCNgb} & $92.36$ & $93.65$ & $95.29$ & $96.04$ & $\mathbf{96.43}$ \\
        \hline
    \end{tabularx}
\end{table}

\subsubsection{Signed Space Origin}
The top part of Table~\ref{tab:ResultsSGCN} shows the results on the \textit{Signed SpaceOrigin} dataset. Increasing the number of layers in the convolutional network results in better performances for all variants, but to different extents. For instance, when going from 1 to 5 layers, the performance gain is $+7,1$ $F$-measure point for \texttt{WSGCNgb} but only $+0.11$ for \texttt{WSGCN±}. With an average diameter of $5.47$ in this dataset, using 5 layers allows for almost complete graph coverage.

All three variants that ignore any form of balance (\texttt{WSGCN+}, \texttt{WSGCN-}, and \texttt{WSGCN±}) are below the unsigned baseline (\texttt{SGCN}), which indicates that using signs improperly is counterproductive as it decreases the classification performance. \texttt{WSGCN+} largely outperforms both other variants, probably because there are many more positive than negative edges in this dataset (cf. Table~\ref{tab:StatsDatasets}).

Among the variants that take balance into account, \texttt{WSGCNgb} consistently outperforms \texttt{WSGCNsb}: $55.67$ vs. $73.69$ with five layers. This shows that the type of balance selected when learning the representation must match the structural properties of the considered graphs. Interestingly, vanilla \texttt{SGCN} is the second-best method, which illustrates the methodological importance of the interconnection scheme when using a master node approach. It is on par with \texttt{WSGCNgb} when using a single layer because they are equivalent for this specific parameter value, however, the difference quickly grows with the number of layers.

\subsubsection{Correlation Clustering Solutions}
The middle part of Table~\ref{tab:ResultsSGCN} shows the $F$-measure scores for the \textit{Correlation Clustering Solutions}. On this dataset, all methods yield quite similar results, except for \texttt{WSGCNgb}, which once again obtains the best performances. In particular, this method is able to capture more information at the whole-graph level than \texttt{SGCN} at the vertex level. Increasing the number of layers still improves the results, but the effect is much weaker than for the previous dataset. The graphs are smaller there, which may explain this, as more layers do not bring more information after a certain point. These results also show that this task is not as challenging as assumed when discussing \textit{Graph2vec} results, since it is possible to get scores much higher than the expected performance of a random classifier. There is still room for improvement, though, as we are far from a perfect classification.

\subsubsection{European Parliament Factions}
The bottom part of Table~\ref{tab:ResultsSGCN} shows the results obtained for the EPF graphs. The behavior on this dataset is similar enough to that on \textit{Signed SpaceOrigin}: the three variants that ignore balance (\texttt{WSGCN+}, \texttt{WSGCN-}, and \texttt{WSGCN±}) are below the original \texttt{SGCN} method (the latter being the worst, again), whereas \texttt{WSGCNgb} gets the best results, peaking at $96.43\%$ when using 5 layers. There are three differences, though: First, the overall performance is much better, with a minimal $F$-measure of $88.45$. This is in line with the behavior exhibited by the Graph2vec variants on the same dataset, and could be explained by the low level of Frustration (around $0.20$). Second, \texttt{WSGCN+} and \texttt{WSGCN-} perform very similarly. Third, \texttt{WSGCNsb} is above \texttt{SGCN}, which could mean that many graphs have a 2-cluster structure in this dataset.

\subsection{Comparison and Concluding Remarks}
\label{sec:ResComp}
We now compare and analyze the results of the three families of methods. Our baseline, which relies on SiNE, is consistently the least efficient method for all tasks. The gap in $F$-measure reaches up to $27.23$ points with the best method, on the \textit{Signed SpaceOrigin} dataset. On the one hand, this could be explained in part by the order and size of the graphs that constitute our benchmark: these are relatively small, whereas SiNE was designed to handle large graphs, with hundreds or thousands of vertices. On the other hand, one could assume that working with smaller graphs should be an advantage when aggregating vertex representations to produce a whole-graph representation, as the most important vertices are likely to have a stronger effect on the produced representation. Nevertheless, our results indicate that applying a method designed to handle whole graphs directly leads to much better classification results than simply aggregating multiple vertex representations. 

Signed Graph2vec, through its \texttt{SG2Vsb} variant, obtains the best performance on the \textit{Signed SpaceOrigin} dataset, while Whole-graph SGCN, through its \texttt{WSGCNgb} variant, largely dominates on both other datasets (and is close on SPO). Regarding the use of signs to learn whole-graph representations, we identify three main results. First, the best signed methods systematically dominate their unsigned counterparts, often by a large margin. This shows the interest of leveraging this information to produce a relevant representation of signed graphs. Second, among the signed methods, those based on some form of SB, be it strict (\texttt{SG2Vsb} and \texttt{WSGCNsb}) or generalized (\texttt{WSGCNgb}), obtain better results than signed methods that ignore this property (\texttt{SG2Vn}, \texttt{WSGCN+}, \texttt{WSGCN-}, \texttt{WSGCN}\texttt{±}). Moreover, the latter generally gets a performance comparable to unsigned methods. This confirms that using signs is not sufficient: the notion of structural balance should be integrated in the design of the representation learning method, so that this property is preserved in the representation space. Third, the generalized version of structural balance seems to work better than the strict version. This is probably because GB is the most general of the two definitions, and subsumes SB. In other words, if the graph exhibits strict structural balance, it is captured by the GB-based representations, whereas SB-based ones cannot handle generalized balance. 

Increasing the number of iterations in G2V and SG2V, or that of layers in SGCN and WSGCN, has a positive impact on performances for all datasets, and for almost all variants. This effect is generally stronger when the appropriate type of balance is leveraged to aggregate the representation of direct and indirect neighbors, though, which confirms the importance of this concept when dealing with signed graphs. For the WSGCN variants, it also shows that the method does not suffer from oversmoothing on the considered datasets. However, at some point, using more iterations or layers does not bring any significant performance gain. 

Graph Convolutional Networks are the best-performing method, overall. However, this comes at a cost: they are also the most expensive in terms of computational runtime. Learning the representation of a graph with WSGCN variants takes more than $25$ seconds on average, over the three datasets. As a comparison, SG2V variants take an average of $0.15$ seconds per graph, and the SiNE baseline takes $0.65$ seconds. SG2V is more than 100 times faster than WSGCN, but its performance is 8 F-measure points below, overall. 

The time complexity of all three methods is linear in $|\mathcal{G}|$, the number of graphs in the collection, so they are likely to scale well on larger datasets. Their complexity is also linear in the number of vertices $N$, therefore, they can reasonably be expected to handle larger graphs as well. The complexity of SiNE and SG2V is independent of the number of edges $M$, whereas for WSGCN, it depends linearly on this graph size. This can be an issue if dealing with much denser graphs, but, as already explained in Section~\ref{sec:BackWholeGrEmb}, real-world networks are typically sparse. The complexity of both SG2V and WSGCN depends on $T$, a variable that controls the extent of the vertex neighborhood considered when building the graph representation. As a consequence, dealing with larger graphs may require increasing $T$ to produce relevant representations. On this point, SWGCN has an advantage over SG2V, as its complexity only depends linearly on $T$, whereas this dependence is quadratic for SG2V. 


\section{Conclusion}
\label{sec:Conclusion}
In this paper, we tackle the problem of learning signed whole-graph representations, and use them for the classification of signed networks. In the absence of any appropriate method in the literature, we generalize two existing models: 1) we adapt Graph2vec to signed graphs by generalizing the Weisfeiler--Lehman relabeling procedure in two different ways; 2) we adapt Signed Graph Convolutional Networks to whole graphs by introducing master nodes and five interconnection schemes. Their implementation is shared online\footref{ftn:Software}. We build a benchmark of signed graphs annotated for classification, which is publicly available online\footref{ftn:Datasets}. It is constituted of three datasets, each one corresponding to a real-world application of graph classification: detecting abusive behavior in online conversations; estimating the multiplicity of optimal solutions to a combinatorial problem; and predicting the polarity at the European Parliament. We use this benchmark to assess our methods and compare them with a baseline relying on SiNE, a standard signed vertex embedding method. Our results show that a signed whole-graph embedding manages to learn better representations for the classification of signed networks. Our proposed method, \texttt{WSGCNgb}, which relies on the notion of generalized structural balance, obtains the best results, overall. 

Our work could be extended in several ways. First, we want to expand our benchmark by including new collections containing larger graphs. This could be done by complementing existing unsigned graph datasets, as we did for the \textit{Signed Space Origin} collection. Repositories of correlation matrices constitute another promising source, as these can be considered signed adjacency matrices. Regarding the methods, we want to apply the master node approach to other types of signed GNN, such as Graph Attention Networks (GATs). It would also be interesting to generalize our methods so that they can use edge weights, which are available in two of the original datasets used in this article. Another promising perspective is to experiment with other master node interconnection schemes, based on other variants of Structural Balance. The concept of \textit{Relaxed Balance}~\cite{Doreian2009, Figueiredo2013}, in particular, is very interesting, as it allows inter-cluster (resp. intra-cluster) edges to be positive (resp. negative).

\bibliographystyle{plainnat}
\bibliography{Cecillon2023_ieee.bib}

\appendix
\section{Additional Results}
\label{sec:AddPerf}
This appendix provides additional results regarding the experiments conducted in Section~\ref{sec:Results}. Tables~\ref{tab:ResultsPreSiNE} and~\ref{tab:ResultsRecSiNE} show the performance of the SiNE baseline, similarly to Table~\ref{tab:ResultsSiNE}, but in terms of Precision and Recall instead of $F$-measure, respectively. Likewise, Tables~\ref{tab:ResultsPreG2V} and~\ref{tab:ResultsRecG2V} show the performance of the G2V-based methods, similarly to Table~\ref{tab:ResultsG2V}, but in terms of Precision and Recall. Finally, Tables~\ref{tab:ResultsPreSGCN} and~\ref{tab:ResultsRecSGCN} show the performance of the SGCN-based methods, similarly to Table~\ref{tab:ResultsSGCN}, but in terms of Precision and Recall. The pairs of matching Precision and Recall scores are very similar. As a result, the $F$-measure scores discussed in the main text are also very similar, since the $F$-measure is the harmonic mean of Precision and Recall. 

\begin{table}[!htb]
    \caption{Results in terms of macro Precision obtained with \texttt{SiNE}. Each row focuses on a task, whereas columns represent the function used to aggregate vertex representations.}
    \label{tab:ResultsPreSiNE}
    \centering
    \begin{tabularx}{\linewidth}{X r r}
        \hline
        \textbf{Task} & \textbf{Sum} & \textbf{Average} \\
        \hline
        SSO & 56.12 & 51.01 \\
        CCS & 50.88 & 49.11 \\
        EPF & 69.07 & 66.74 \\
        \hline
    \end{tabularx}
\end{table}

\begin{table}[!htb]
    \caption{Results in terms of macro Recall obtained with \texttt{SiNE}. Each row focuses on a task, whereas columns represent the function used to aggregate vertex representations.}
    \label{tab:ResultsRecSiNE}
    \centering
    \begin{tabularx}{\linewidth}{X r r}
        \hline
        \textbf{Task} & \textbf{Sum} & \textbf{Average} \\
        \hline
        SSO & 54.74 & 49.44 \\
        CCS & 50.09 & 48.14 \\
        EPF & 70.57 & 68.45 \\
        \hline
    \end{tabularx}
\end{table}

\begin{table}[!htb]
    \caption{Results in terms of macro Precision obtained with \texttt{Graph2vec} and our two proposed signed adaptations. Each column focuses on a specific number of iterations used when extracting rooted subgraphs.}
    \label{tab:ResultsPreG2V}
    
    \centering
    \begin{tabularx}{\linewidth}{X l r r r r r r r}
        \hline
        \textbf{Task} & \textbf{Method} & \textbf{1 it.} & \textbf{2 it.} & \textbf{3 it.} & \textbf{4 it.} & \textbf{5 it.} \\
        \hline
        SSO & \texttt{G2V} & 76.31 & 71.49 & 73.01 & 73.94 & 74.03 \\
        & \texttt{SG2Vn} & 76.02 & 72.20 & 72.89 & 73.49 & 72.98 \\
        & \texttt{SG2Vsb} & 66.68 & 71.58 & 74.79 & 76.50 & 77.03 \\
        \hline
        CCS & \texttt{G2V} & 48.67 & 47.97 & 46.21 & 52.87 & 51.79 \\
        & \texttt{SG2Vn} & 50.33 & 49.88 & 48.53 & 53.20 & 51.84 \\
        & \texttt{SG2Vsb} & 50.31 & 52.13 & 49.67 & 51.49 & 52.02 \\
        \hline
        EPF & \texttt{G2V} & 47.68 & 51.24 & 54.53 & 61.85 & 64.78 \\
        & \texttt{SG2Vn} & 74.75 & 81.05 & 84.29 & 86.35 & 88.13 \\
        & \texttt{SG2Vsb} & 79.66 & 82.85 & 86.46 & 87.64 & 90.44 \\
        \hline
    \end{tabularx}
\end{table}

\begin{table}[!htb]
    \caption{Results in terms of macro Recall obtained with \texttt{Graph2vec} and our two proposed signed adaptations. Each column focuses on a specific number of iterations used when extracting rooted subgraphs.}
    \label{tab:ResultsRecG2V}
    
    \centering
    \begin{tabularx}{\linewidth}{X l r r r r r r r}
        \hline
        \textbf{Task} & \textbf{Method} & \textbf{1 it.} & \textbf{2 it.} & \textbf{3 it.} & \textbf{4 it.} & \textbf{5 it.} \\
        \hline
        SSO & \texttt{G2V} & 73.90 & 71.15 & 72.23 & 73.98 & 73.51 \\
        & \texttt{SG2Vn} & 73.71 & 70.69 & 71.42 & 72.28 & 71.77 \\
        & \texttt{SG2Vsb} & 67.91 & 72.44 & 74.97 & 77.46 & 77.85 \\
        \hline
        CCS & \texttt{G2V} & 48.19 & 46.40 & 45.73 & 50.76 & 52.35 \\
        & \texttt{SG2Vn} & 49.08 & 48.63 & 48.21 & 51.95 & 52.40 \\
        & \texttt{SG2Vsb} & 49.37 & 51.49 & 49.73 & 51.25 & 51.22 \\
        \hline
        EPF & \texttt{G2V} & 43.74 & 48.36 & 50.26 & 59.07 & 61.58 \\
        & \texttt{SG2Vn} & 76.53 & 81.83 & 84.07 & 85.97 & 89.23 \\
        & \texttt{SG2Vsb} & 82.03 & 82.95 & 86.16 & 88.34 & 89.52 \\
        \hline
    \end{tabularx}
\end{table}

\begin{table}[!t]
    \caption{Results in terms of macro Precision obtained with the original \texttt{SGCN} and our five proposed \texttt{WSGCN} interconnection schemes. Each column focuses on a specific number of convolution layers.}
    \label{tab:ResultsPreSGCN}
    
    \centering
    \begin{tabularx}{\linewidth}{X l r r r r r}
        \hline
        \textbf{Task} & \textbf{Method} & \textbf{1 lay.} & \textbf{2 lay.} & \textbf{3 lay.} & \textbf{4 lay.} & \textbf{5 lay.} \\
        \hline
        SSO & \texttt{SGCN} & 65.88 & 66.76 & 68.02 & 68.45 & 69.40 \\
        & \texttt{WSGCN+} & 65.24 & 66.98 & 68.57 & 69.68 & 70.00 \\
        & \texttt{WSGCN-} & 54.42 & 55.11 & 55.78 & 55.48 & 56.03 \\
        & \texttt{WSGCN±} & 49.84 & 48.99 & 49.25 & 49.18 & 49.58 \\
        & \texttt{WSGCNsb} & 53.16 & 55.84 & 57.10 & 56.42 & 56.59 \\
        & \texttt{WSGCNgb} & 68.32 & 68.59 & 72.35 & 72.46 & 73.51 \\
        \hline
        CCS & \texttt{SGCN} & 71.04 & 71.21 & 71.86 & 72.01 & 72.07 \\
        & \texttt{WSGCN+} & 71.25 & 71.49 & 71.67 & 72.23 & 71.77 \\
        & \texttt{WSGCN-} & 70.42 & 70.41 & 70.76 & 71.04 & 71.40 \\
        & \texttt{WSGCN±} & 70.99 & 71.45 & 71.20 & 71.43 & 71.33 \\
        & \texttt{WSGCNsb} & 69.84 & 70.33 & 71.25 & 71.49 & 71.94 \\
        & \texttt{WSGCNgb} & 71.43 & 72.35 & 73.56 & 74.03 & 74.18 \\
        \hline

        EPF & \texttt{SGCN} & 90.46 & 91.20 & 91.89 & 92.08 & 92.58 \\
        & \texttt{WSGCN+} & 89.40 & 89.96 & 90.48 & 91.24 & 92.23 \\
        & \texttt{WSGCN-} & 88.95 & 89.39 & 90.66 & 91.54 & 92.03 \\
        & \texttt{WSGCN±} & 86.11 & 88.23 & 89.86 & 90.67 & 91.60 \\
        & \texttt{WSGCNsb} & 91.32 & 92.20 & 93.56 & 94.67 & 95.45 \\
        & \texttt{WSGCNgb} & 92.66 & 93.81 & 95.45 & 96.16 & 96.51 \\
        \hline
    \end{tabularx}
\end{table}

\begin{table}[!t]
    \caption{Results in terms of macro Recall obtained with the original \texttt{SGCN} and our five proposed \texttt{WSGCN} interconnection schemes. Each column focuses on a specific number of convolution layers.}
    \label{tab:ResultsRecSGCN}
    
    \centering
    \begin{tabularx}{\linewidth}{X l r r r r r}
        \hline
        \textbf{Task} & \textbf{Method} & \textbf{1 lay.} & \textbf{2 lay.} & \textbf{3 lay.} & \textbf{4 lay.} & \textbf{5 lay.} \\
        \hline
        SSO & \texttt{SGCN} & 67.09 & 67.66 & 68.10 & 69.29 & 69.68 \\
        & \texttt{WSGCN+} & 65.00 & 66.58 & 68.27 & 68.29 & 68.59 \\
        & \texttt{WSGCN-} & 53.96 & 54.67 & 55.34 & 55.70 & 55.75 \\
        & \texttt{WSGCN±} & 48.73 & 48.91 & 48.77 & 49.32 & 49.20 \\
        & \texttt{WSGCNsb} & 52.22 & 54.16 & 53.96 & 53.80 & 54.77 \\
        & \texttt{WSGCNgb} & 64.94 & 69.11 & 70.10 & 72.10 & 73.87 \\
        \hline
        CCS & \texttt{SGCN} & 69.55 & 70.09 & 70.68 & 70.85 & 71.71 \\
        & \texttt{WSGCN+} & 69.76 & 70.24 & 70.59 & 70.33 & 70.93 \\
        & \texttt{WSGCN-} & 69.86 & 70.69 & 70.92 & 71.00 & 70.96 \\
        & \texttt{WSGCN±} & 70.53 & 70.59 & 70.80 & 70.97 & 71.41 \\
        & \texttt{WSGCNsb} & 69.34 & 69.91 & 70.47 & 70.93 & 70.98 \\
        & \texttt{WSGCNgb} & 72.07 & 72.05 & 72.40 & 72.46 & 72.81 \\
        \hline
        EPF & \texttt{SGCN} & 89.86 & 90.54 & 91.37 & 92.00 & 92.72 \\
        & \texttt{WSGCN+} & 87.73 & 88.64 & 90.50 & 90.98 & 91.51 \\
        & \texttt{WSGCN-} & 88.27 & 89.21 & 90.32 & 90.64 & 91.57 \\
        & \texttt{WSGCN±} & 86.53 & 88.67 & 90.16 & 90.89 & 91.52 \\
        & \texttt{WSGCNsb} & 90.90 & 91.98 & 93.06 & 93.67 & 94.53 \\
        & \texttt{WSGCNgb} & 92.06 & 93.49 & 95.13 & 95.92 & 96.35 \\
        \hline
    \end{tabularx}
\end{table}

\end{document}